\documentclass[journal]{IEEEtai}

\usepackage{color,array}

\usepackage{graphicx}
\usepackage{ulem}
\usepackage{graphicx}
\usepackage{comment}
\usepackage{amsmath,amssymb} 
\usepackage{color}

\usepackage{url}
\usepackage{booktabs}
\usepackage{nicefrac}
\usepackage{microtype}
\usepackage{epsfig}
\usepackage[tight,footnotesize]{subfigure}
\usepackage{amsmath}
\usepackage{amssymb}
\usepackage{amsthm}
\usepackage{bm}
\usepackage{multirow}
\usepackage{makecell}
\usepackage{footmisc}
\usepackage{marvosym}
\usepackage{latexsym}
\usepackage[utf8]{inputenc}
\usepackage{kotex}
\usepackage{titlesec}
\usepackage{caption}
\usepackage{algorithm,algpseudocode}
\usepackage{tablefootnote}

\newcommand{\etal}{\textit{et al}.}
\newcommand{\ie}{\textit{i}.\textit{e}., }
\newcommand{\eg}{\textit{e}.\textit{g}., }

\usepackage{dsfont}
\usepackage{pifont}
\usepackage[bookmarksnumbered=true]{hyperref} 
\hypersetup{
     colorlinks = true,
     linkcolor = blue,
     anchorcolor = blue,
     citecolor = blue,
     filecolor = blue,
     urlcolor = blue
     }

\begin{document}

\title{Beyond Classification: Directly  Training Spiking Neural Networks for Semantic Segmentation} 

\author{Youngeun Kim, Joshua Chough, and Priyadarshini Panda \thanks{ Youngeun Kim, Joshua Chough, and Priyadarshini~Panda are with the Department of Electrical Engineering, Yale University, New Haven, CT, USA.
 (Youngeun Kim and Joshua Chough contributed equally to this work.)
(Corresponding author: Youngeun Kim; E-mail: youngeun.kim@yale.edu).}
\\Department of Electrical Engineering, Yale University, USA
}


\maketitle

\begin{abstract}
Spiking Neural Networks (SNNs) have recently emerged as the low-power alternative to Artificial Neural Networks (ANNs) because of their sparse, asynchronous, and binary event-driven processing.
Due to their energy efficiency, SNNs have a high possibility of being deployed for real-world, resource-constrained systems such as autonomous vehicles and drones.
However, owing to their non-differentiable and complex neuronal dynamics, most previous SNN optimization methods have been limited to image recognition.
In this paper, we explore the SNN applications beyond classification and present semantic segmentation networks configured with spiking neurons.
Specifically, we first investigate two representative SNN optimization techniques for recognition tasks (\ie ANN-SNN conversion and surrogate gradient learning) on semantic segmentation datasets.
We observe that, when converted from ANNs, SNNs suffer from high latency and low performance due to the spatial variance of features.
Therefore, we directly train networks with surrogate gradient learning, resulting in lower latency and higher performance than ANN-SNN conversion.
Moreover, we redesign two fundamental ANN segmentation architectures (\ie Fully Convolutional Networks and DeepLab) for the SNN domain.
We conduct experiments on two public semantic segmentation benchmarks including the PASCAL VOC2012 dataset and the DDD17 event-based dataset.
In addition to showing the feasibility of SNNs for semantic segmentation, we show that SNNs can be more robust and energy-efficient compared to their ANN counterparts in this domain. 
\end{abstract}



\section{Introduction}

\IEEEPARstart{A}{rtificial} Neural Networks (ANNs) have shown impressive performance across various computer vision fields for a few decades \cite{he2016deep,simonyan2014very,girshick2015fast}. 
However,
ANNs suffer from huge computational costs \cite{sze2017efficient},
limiting their application in power-hungry systems such as Internet-of-Things (IoT) devices.
As an alternative to low-power ANNs, recent studies have focused on bio-plausible Spiking Neural Networks (SNNs) \cite{roy2019towards,deng2020rethinking}, which process visual information with temporal binary events (\ie spikes).
Spikes stimulate neuronal membrane potentials and convey discriminative information from shallow layers to deep layers.
SNN neurons only consume energy whenever spikes are generated which allow for these asynchronous processes to be implemented on highly energy-efficient neuromorphic hardware \cite{furber2014spinnaker,akopyan2015truenorth,davies2018loihi}.

\begin{figure}[t]
\begin{center}
\def\arraystretch{0.5}
\begin{tabular}{@{}c@{\hskip 0.05\linewidth}c@{}c}
\includegraphics[width=0.9\linewidth]{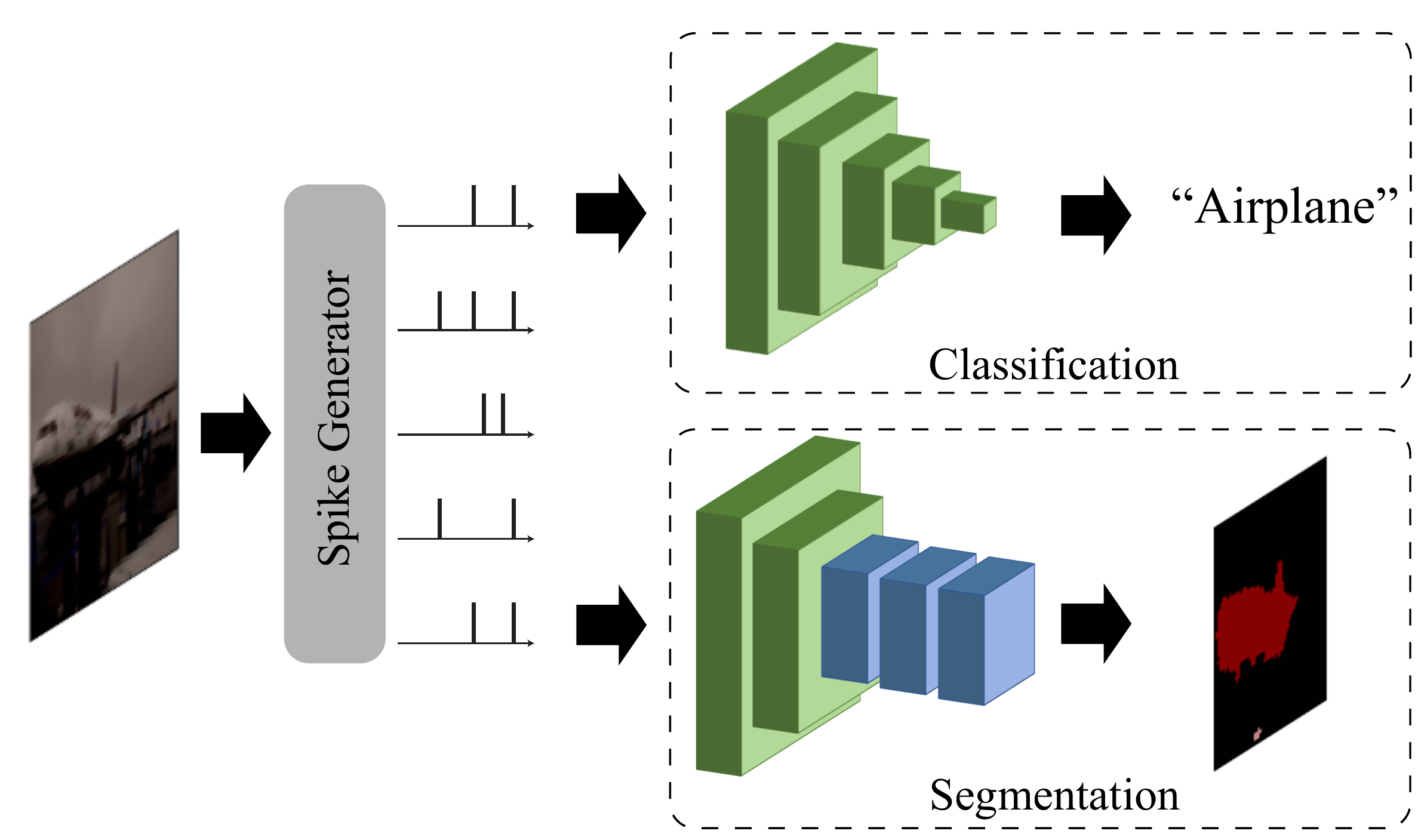}  \\
\end{tabular}
\end{center}
\caption{The concept of image classification and semantic segmentation. For a given image, the classification network  provides a class prediction (\eg Airplane). On the other hand, the semantic segmentation network maintains the high resolution feature map at the end of the network (colored in blue) and  assigns every pixel in the image its own class, resulting in a two-dimensional prediction map.
}
\label{fig:intro:class_seg_concept}
\end{figure}

In order to exploit the energy advantage of binary information transmission, optimization algorithms for SNNs have been developed in the past few decades, with a focus on image classification.
Among various training algorithms, the ANN-SNN conversion method \cite{sengupta2019going,han2020rmp,diehl2015fast,rueckauer2017conversion} has been highlighted as a result of its simplicity and high performance.
Conversion replaces the neurons of a pre-trained ANN using the ReLU activation function with Integrate-and-Fire (IF) neurons in an SNN.
The conversion method avoids the complexity of spike-based training since it relies on ANN training and thus, yields high performance on complex data.
%
%
However, since conversion naively scales the firing thresholds with the maximum layer activations, it is hard to capture spike information across the spatial and temporal domain.
As a result, the SNN created using ANN-SNN conversion requires thousands of time-steps to achieve similar performance as the original pre-trained ANN, and has been limited to fundamental vision tasks such as image classification.

Considering the likelihood of the future deployment of SNNs in scene-understanding tasks, it is essential to investigate the applicability of SNNs for beyond classification.
For example, a vision system for autonomous vehicles should contain an energy-efficient neural module for analyzing the overall context of scenery \cite{yurtsever2020survey,treml2016speeding}.
Therefore, in this paper, we move beyond classification by exploring semantic segmentation for neuromorphic systems.
As shown in Fig. \ref{fig:intro:class_seg_concept}, a semantic segmentation network conducts pixel-wise classification, resulting in a two-dimensional prediction map.
Since partitioning multiple foreground objects from the background is an essential and fundamental vision task, semantic segmentation has been extensively studied in the ANN domain \cite{he2016deep,kim2019cnn,zhang2018context,chen2018encoder,long2015fully}.
Unfortunately, we found that the state-of-the-art ANN-SNN conversion technique fails to achieve reasonable SNN performance on a semantic segmentation dataset.
This is because segmentation datasets cause ANNs to have huge activation variance across the spatial and channel axes (will be discussed in Section \ref{section:preliminary study}).
Moreover, ANN-SNN conversion cannot be compatible with a spike stream dataset from Dynamic Vision Sensor (DVS) camera \cite{patrick2008128x, lichtsteiner2006128, posch2014retinomorphic, delbruck2010activity}, which limits the potential advantage of a neuromorphic system.



To address the aforementioned problems, we focus on surrogate gradient backpropagation learning \cite{neftci2019surrogate,lee2016training,wu2018spatio} rather than ANN-SNN conversion, which directly trains SNNs from input spikes.
The major difficulty of gradient backpropagation learning with SNNs is the non-differentiable neuronal functionality of SNNs.
A Leaky Integrate-and-Fire (LIF) neuron does not generate spikes before its membrane potential exceeds a firing threshold, resulting in a non-differentiable point during backpropagation.
Therefore, we approximate the backward gradient function of an LIF neuron using a piece-wise linear function during backpropagation.
Using the approximated gradient function, the weight parameters are optimized in order to minimize spatial cross-entropy loss.
In addition, we leverage Batch Normalization Through Time (BNTT) \cite{kim2020revisiting}, a recent work proposes a temporal batch normalization, which enables our segmentation networks to be trained from scratch. 
As a result, SNNs with BNTT-backpropagation can learn the temporal dynamics of input spikes, with shorter number of time-steps and better performance compared to ANN-SNN conversion.

With surrogate learning, we investigate two representative segmentation architectures from the ANN domain, \ie  Fully Convolutional Networks (FCNs) \cite{long2015fully} and 
DeepLab with dilated (\ie atrous) convolutional layers \cite{he2016deep}.
The FCN architecture consists of an encoder-decoder architecture, which recovers the resolution of the original image with the upsampling decoder.
On the other hand, DeepLab only has an encoder architecture but uses dilated convolutions to cover wide receptive fields without computational overhead.
Due to the fact that very deep SNN architectures suffer performance degradation because of the mismatch between real gradients and approximated gradients, we use the VGG9 architecture as the backbone for our networks for segmentation.



In summary, the main contributions of this work are as follows:
(i) To the best of our knowledge, our work is the first work to train SNNs for semantic segmentation.
This is an important research direction given that low-power SNNs will be deployed in scene-understanding tasks.
(ii) We investigate two representative SNN optimization techniques (\eg ANN-SNN conversion and surrogate gradient learning) on semantic segmentation datasets.
We observe that surrogate gradient learning achieves better performance with lower latency compared to ANN-SNN conversion.
(iii)
We present Spiking-FCN and Spiking-DeepLab, which expand upon the two fundamental architectures used for semantic segmentation in the ANN domain.
Also, we conduct comprehensive experiments on various benchmarks including static PASCAL VOC2012 and DDD17 from a DVS camera.
The proposed segmentation architectures can bring a more than $2\times$ energy efficiency than standard ANNs.

The paper is organized as follows. Section II introduces background knowledge on SNNs and the semantic segmentation work done in the ANN domain.
Section III presents our experiments on ANN-SNN conversion.
Section IV details our methods of training segmentation networks with spiking neurons.
Lastly, in Section V, we present our comprehensive experimental results on both the static and video segmentation datasets.
At the end of this paper, we provide our conclusions and point out future work directions.

\begin{figure}[t!]
  \begin{center}
    \includegraphics[width=0.5\textwidth]{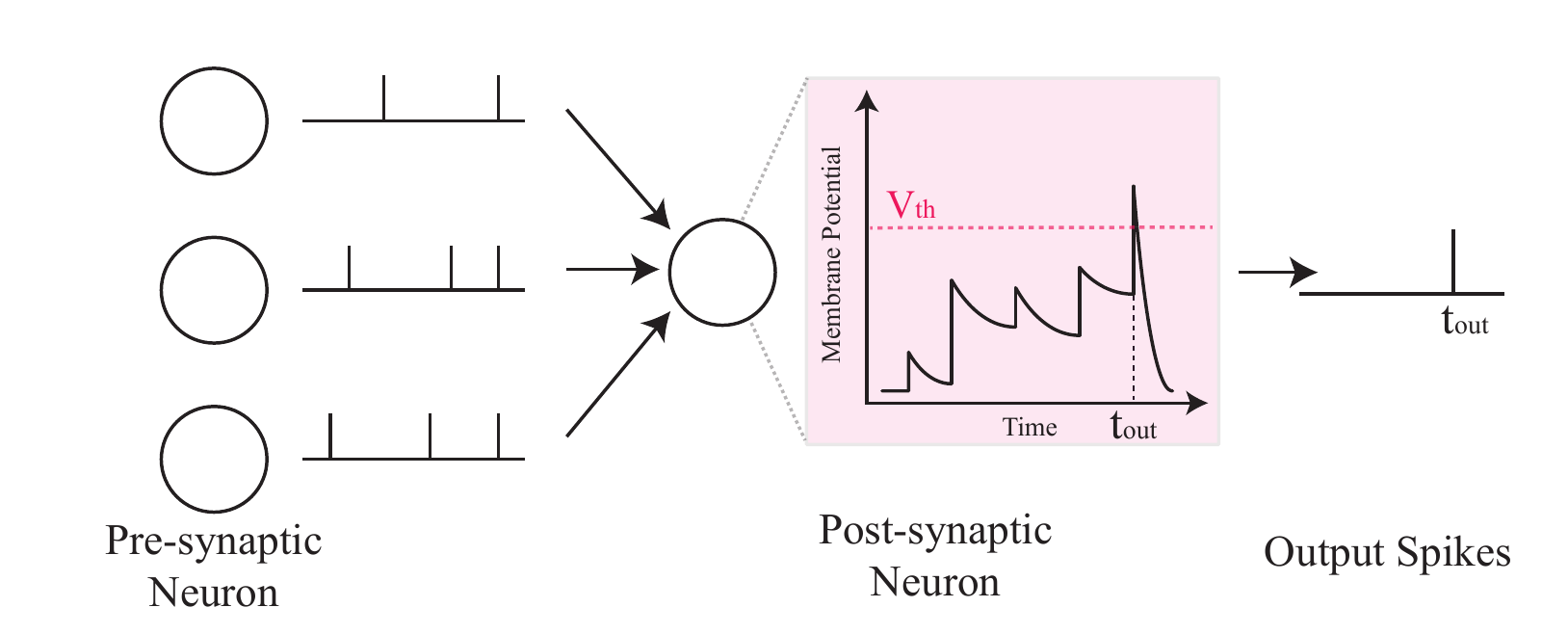}
  \end{center}
  \caption{ The neuronal dynamics of SNNs. Pre-synaptic neurons convey spike trains to a post-synaptic neuron. This increases the membrane potential voltage and the post-synaptic neuron generates the output spikes whenever the membrane potential exceeds the neuronal firing threshold. After that, the membrane potential is set to a reset voltage.  }
   \label{fig:relatedwork:neuron_dynamic}
\end{figure}

\section{Related Work}

\subsection{Spiking Neural Network}
Spiking Neural Networks (SNNs) have emerged as the next generation of neural networks \cite{roy2019towards,panda2020toward,cao2015spiking,diehl2015unsupervised,comsa2020temporal,venkatesha2021federated,kim2021optimizing,kim2021privatesnn}, because they offer huge energy-efficiency advantage over ANNs.
Different from standard ANNs that make use of float values, SNNs process binary spikes (0 or 1) across multiple time-steps.
SNNs take binary spike trains as an input, which can be obtained from both static RGB images and DVS camera data. 
For static images, various coding schemes have been proposed.
Poisson rate coding generates spikes in which the number of spikes is proportional to the pixel intensity.
Due to its simplicity and high performance, Poisson rate coding has been widely used in previous works \cite{roy2019towards,diehl2015unsupervised,lee2016training}.
Temporal coding allows only one spike per neuron, resulting in energy efficiency from fewer spikes. Here, spike latency is inversely proportional to the pixel intensity \cite{mostafa2017supervised,park2020t2fsnn,comsa2020temporal}.
Thus, bright pixels generate more spike events in earlier time-steps than dark pixels.
However, for DVS camera data, SNNs can be directly trained without any code generator.

In terms of an activation function, SNNs commonly use a Leaky Integrate-and-Fire (LIF) neuron \cite{gerstner2002spiking}.
The LIF neuron (Fig. \ref{fig:relatedwork:neuron_dynamic}) has a membrane potential which can store the temporal information by accumulating pre-synaptic spikes.
The neuron generates a post-synaptic spike whenever the membrane potential exceeds a predefined firing threshold.
This integrate-and-fire behavior is non-differentiable, so SNNs are hard to train with standard backpropagation \cite{neftci2019surrogate}.
To address this limitation, work in the past decade has focused on various training techniques for SNNs.
Spike-timing-dependent plasticity (STDP) learning \cite{bi1998synaptic,hebb2005organization,bliss1993synaptic} is based on the neuroscience observation that weight connections can be reinforced or punished according to the temporal correlation of spikes.
STDP learning can be implemented without a complicated backpropagation module but can only be applied to small-scale tasks due to its locality \cite{yousefzadeh2018practical,jin2010implementing}.
ANN-SNN conversion methods have received attention due to their high performance on complex tasks \cite{sengupta2019going,han2020rmp,diehl2015fast,rueckauer2017conversion}.
In order to approximate ReLU activations with LIF activations, pre-trained ANNs are converted to SNNs using weight (or threshold) balancing techniques.
ANN-SNN conversion requires many time-steps to represent the float values of ANNs with binary spikes.
Surrogate gradient learning addresses the non-differentiability of LIF neurons by defining a surrogate gradient function in a backpropagation process \cite{lee2016training,neftci2019surrogate}.
This training scheme enables SNNs to learn the temporal dynamics of spike trains, resulting in small latency and reasonable performance.
Unfortunately, most SNN optimization algorithms are developed for basic, fundamental vision tasks such as image recognition \cite{lee2016training,kim2020revisiting,diehl2015unsupervised}, visualization \cite{kim2021visual}, and optimization \cite{fang2019swarm,frady2020neuromorphic}.

\begin{figure}[t!]
  \begin{center}
  \includegraphics[width=0.9\linewidth]{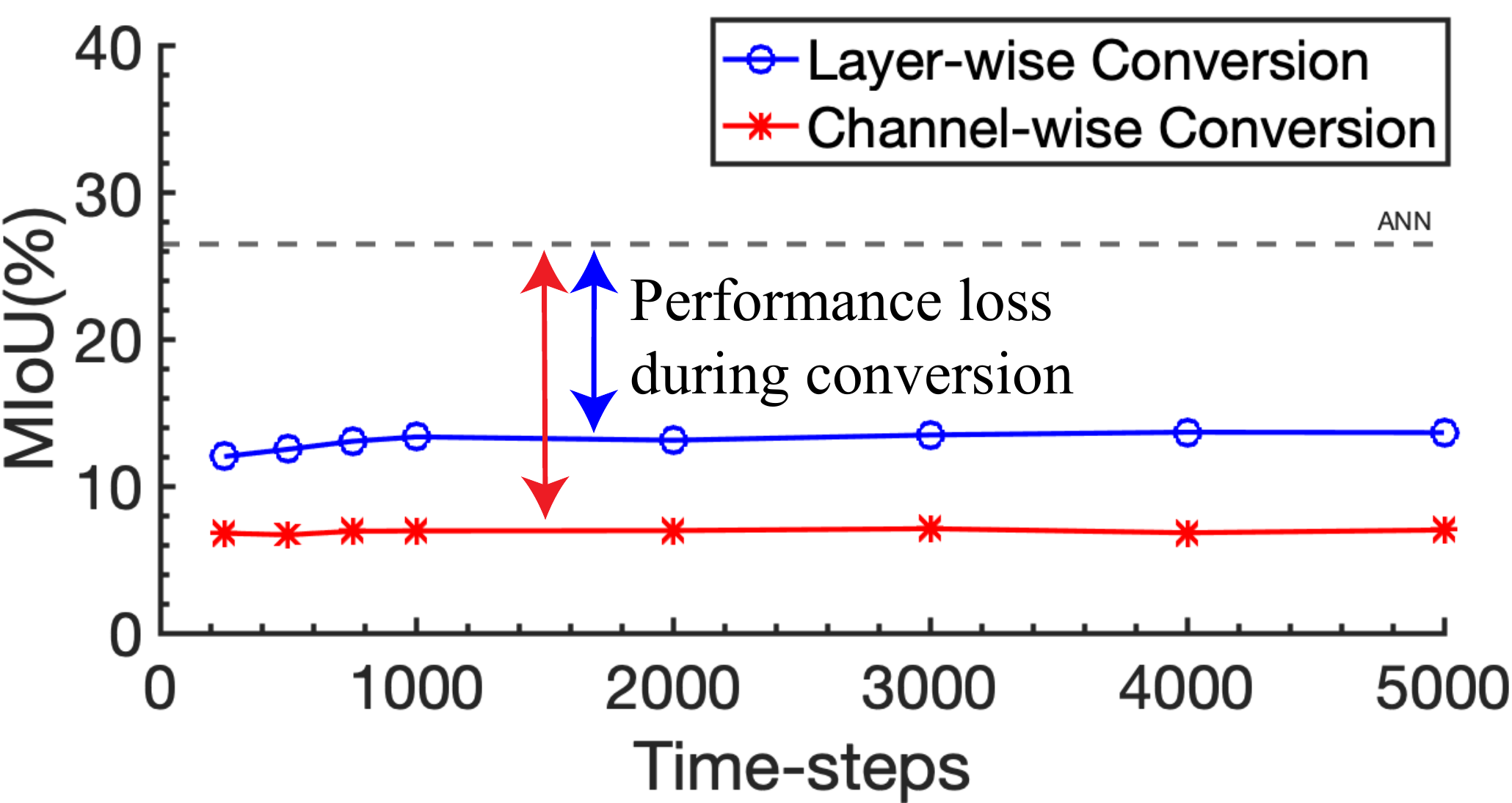} \\
  \end{center}
  \caption{
  {
  Performance of converted SNNs with respect to the number of time-steps. We evaluate two conversion algorithms (\ie layer-wise conversion \cite{sengupta2019going} and channel-wise conversion \cite{kim2020spiking}) with a DeepLab architecture trained on PASCAL VOC2012. Compared to the ANN performance (black dotted line), the converted SNNs suffer from performance degradation caused by a large variation in activations.}
  } 
   \label{fig:prelimin:conversion_studies}
\end{figure}

\begin{figure*}[t]
\begin{center}
\def\arraystretch{0.5}
\begin{tabular}{@{}c@{\hskip 0.05\linewidth}c@{}c}
\includegraphics[width=0.48\linewidth]{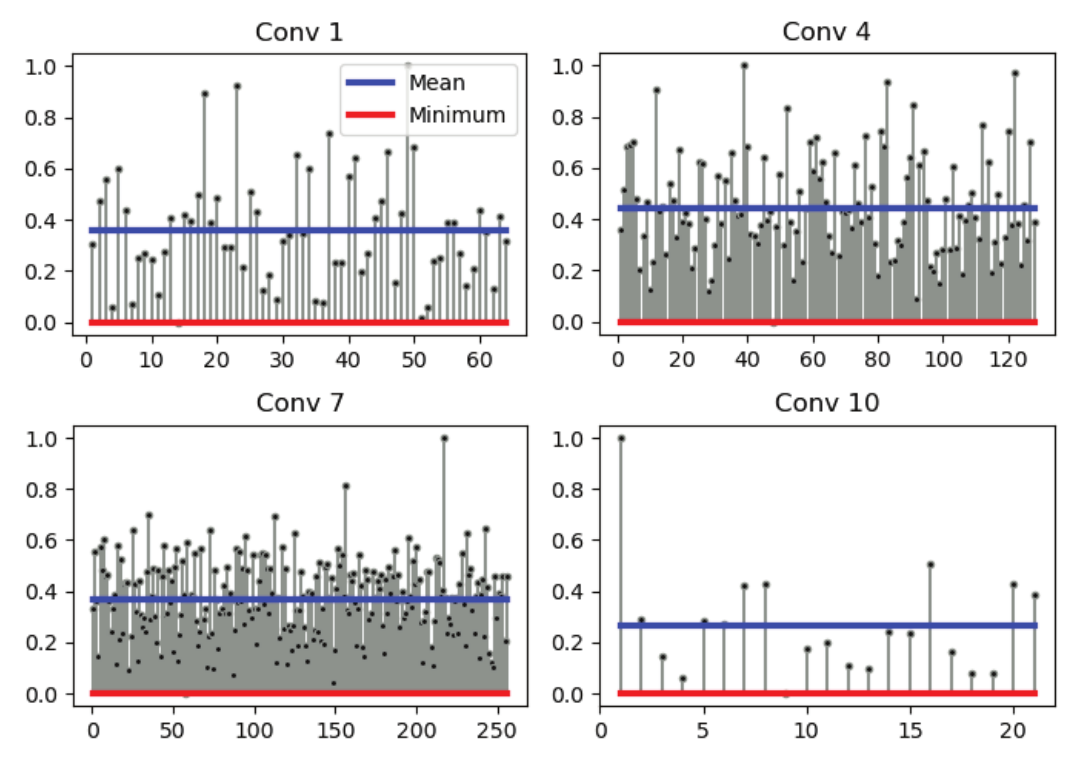} &
\includegraphics[width=0.48\linewidth]{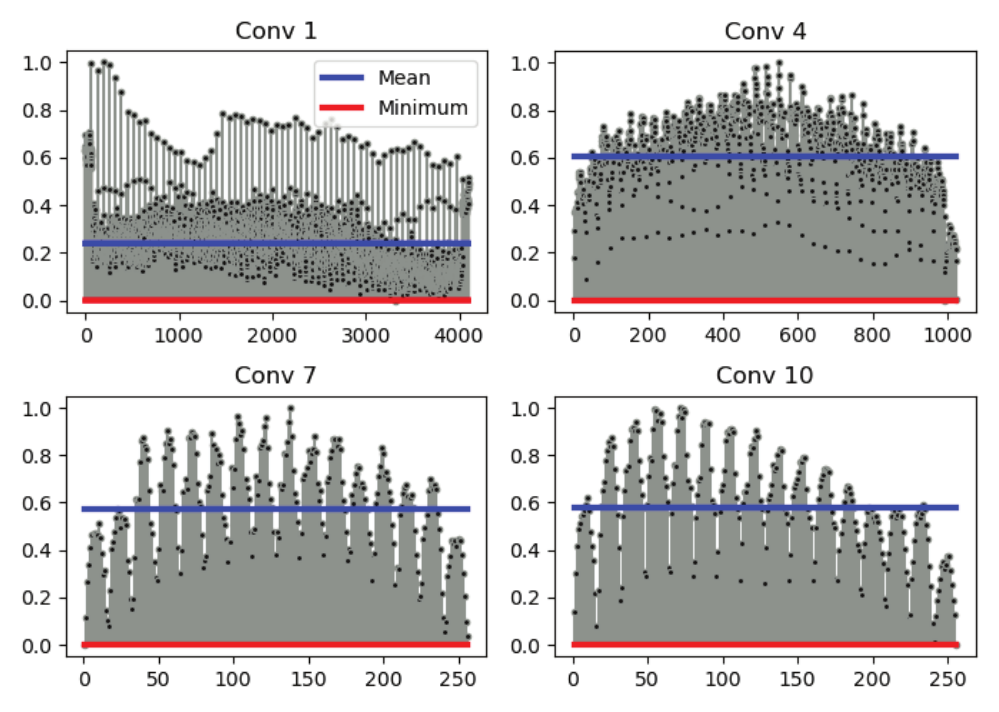} 
\\
{\hspace{4mm} (a) Channel activation  } & {\hspace{4mm} (b)  Spatial activation  }\\
\end{tabular}
\end{center}
\caption{Illustration of normalized maximum activations across  (a) channel and (b) spatial axes. We use a VGG9-based DeepLab architecture trained on PASCAL VOC2012 dataset. In the figure, the \textit{x} and \textit{y} axes denote, respectively, the index of the channel/spatial axes in that layer and the normalized maximum activations across the channel/spatial axes.
}
\label{fig:prelim:spikeactivation}
\end{figure*}

\subsection{Semantic Segmentation}
The objective of semantic segmentation is to classify every pixel in an image with a label, a key task for scene understanding.
This becomes more important with the development of medical image analysis \cite{ronneberger2015u,cciccek20163d}, autonomous vehicles \cite{treml2016speeding,janai2020computer}, and augmented reality \cite{minaee2020image}.
Therefore, semantic segmentation with deep neural networks has been extensively studied in the ANN domain. 
The Fully Convolutional Network (FCN) \cite{long2015fully} is the one of the pioneering works in deep semantic segmentation.
FCNs preserve image details by adding the intermediate high-resolution feature maps into its decoder path.
In a similar way, U-Net \cite{ronneberger2015u} concatenates intermediate feature maps during its up-convolution process.
Different from previous approaches, deconvolution networks \cite{noh2015learning} do not exploit a skip connection but learn a detailed representation from low-resolution feature maps.
RefineNet \cite{lin2017refinenet} uses multi-scale inputs in order to enhance detail in the segmentation map.
DeepLab \cite{he2016deep} utilizes dilated (\ie atrous) convolutions which enlarge the receptive field without any additional computational cost.
Dilated convolutions have become one of the most popular techniques for semantic segmentation since they are easy to be implemented in deep learning frameworks (\eg TensorFlow \cite{tensorflow2015-whitepaper}).
Despite its importance and fast growth in the ANN domain, semantic segmentation has not yet been studied with spiking neurons.
So far, in the SNN domain, a few works \cite{meftah2010segmentation,zheng2019image} have proposed region segmentation without providing semantic information (\ie class information).
In order to fill the gap, in this work, we configure the spiking versions of segmentation networks based on two representative segmentation architectures, \ie FCN and DeepLab.


\section{Preliminary Study}
\label{section:preliminary study}

In this section, we apply ANN-SNN conversion methods to semantic segmentation. 
First, we train the ANN-DeepLab with conventional 2D-cross entropy loss \cite{he2016deep}.
After that, we convert the pre-trained ANN to an SNN with two conversion methods \cite{sengupta2019going,kim2020spiking}.
Sengputa \etal  \cite{sengupta2019going} propose a state-of-the-art conversion technique in the image recognition domain. They take into account spike behavior in the previous layers to achieve accurate weight balancing.
Also, the authors of \cite{kim2020spiking} propose  channel-wise weight balancing for each layer in order to  capture a feature with high variation in an object detection scenario.
We evaluate these methods on a DeepLab architecture  with a VGG9 backbone network \cite{chen2017deeplab} on PASCAL VOC2012 dataset \cite{everingham2010pascal}.
{In Fig. \ref{fig:prelimin:conversion_studies}, we observe that the ANN-SNN conversion process significantly degrades the performance from a pre-trained ANN even with thousands of  time-steps.
This is a relatively huge loss compared to ANN-SNN conversion on image classification which shows less than $1\%$ accuracy degradation \cite{sengupta2019going}.
}
The reason for this is that networks trained for the semantic segmentation task have a large variation of activations in both the channel and spatial axes (Fig. \ref{fig:prelim:spikeactivation}).
This causes ANN-SNN conversion to fail to find a proper scaling factor. 
Thus, applying the same balancing factor across layers (or channels) does not fully preserve delicate activations.
More importantly, ANN-SNN conversion cannot be applied on video spike streams from a DVS camera, since ANN-SNN conversion is only compatible with static images.
Thus, we focus on a direct training approach in order to implement semantic segmentation with spiking neurons.

\section{Methodology}

\subsection{Leaky Integrate-and-Fire (LIF) Neuron}
\label{sec:LIF neuron}
We leverage a Leaky Integrate-and-Fire (LIF) neuron in our spiking segmentation networks. 
Fig. \ref{fig:relatedwork:neuron_dynamic} illustrates the dynamics of an LIF neuron.
The LIF neuron can be represented with a membrane potential and an input signal.
The membrane potential $U_m$ stores the temporal spike information in capacitance. When an input signal $I(t)$ is fed into the LIF neuron, the membrane potential is changed according to the  following differential equation:
\begin{equation}
    \tau_m \frac{dU_m}{dt} = -U_m  + RI(t),
    \label{eq:LIF_origin}
\end{equation}
where, $R$ is an input resistance for the LIF circuit and $\tau_m$ is the time constant for the membrane potential decay.
Since the voltage and current have continuous values, we convert the differential equation into a discrete version in order to conduct digital simulation.
We can represent the membrane potential $u_{i}^{t}$ of a single neuron $i$ at time-step $t$ as: 
\begin{equation}
    u_i^t = \lambda u_i^{t-1} + \sum_j w_{ij}o^t_j.
    \label{eq:LIF}
\end{equation}
Here, the current membrane potential consists of the decayed membrane potential from previous time-steps and the weighted spike signal from the pre-synaptic neurons $j$.
The notation $\lambda$ and $w_{ij}$ are for a leak factor and weight connection between the pre-synaptic neuron $j$ and the post-synaptic neuron $i$, respectively.
When $u_i^{t}$ exceeds the firing threshold, the neuron $i$ generates a spike output $o_i^{t}$:
\begin{equation}
    o^{t}_i =
\begin{cases}
 1,          & \text{if $u_i^{t} >\theta$},  \\
    0
    & \text{otherwise.} 
\end{cases}
\label{eq:firing}
\end{equation}
After the neuron fires, the membrane potential is lowered by the amount of the threshold (\ie soft reset). 
In our experiments, we use a soft reset scheme since it achieves better performance than a hard reset (\ie reset to the minimum voltage such as zero). 
This is because a soft reset allows the neuron to retain residual information after the reset, thus preventing information loss \cite{han2020rmp}.

\begin{figure}[t!]
  \begin{center}
  \includegraphics[width=1.0\linewidth]{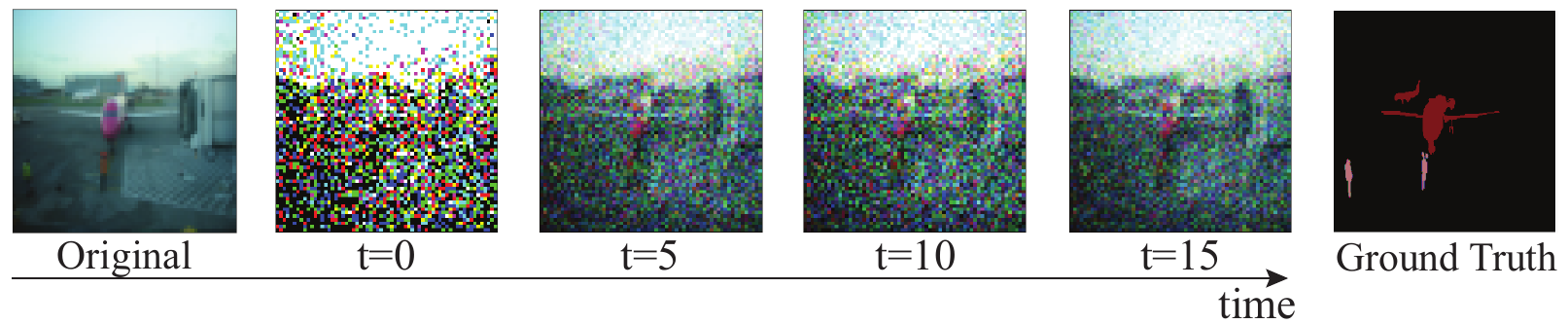} \\
{\hspace{4mm}(a) PASCAL VOC2012} \\ 
\vspace{1mm}
\includegraphics[width=1.0\linewidth]{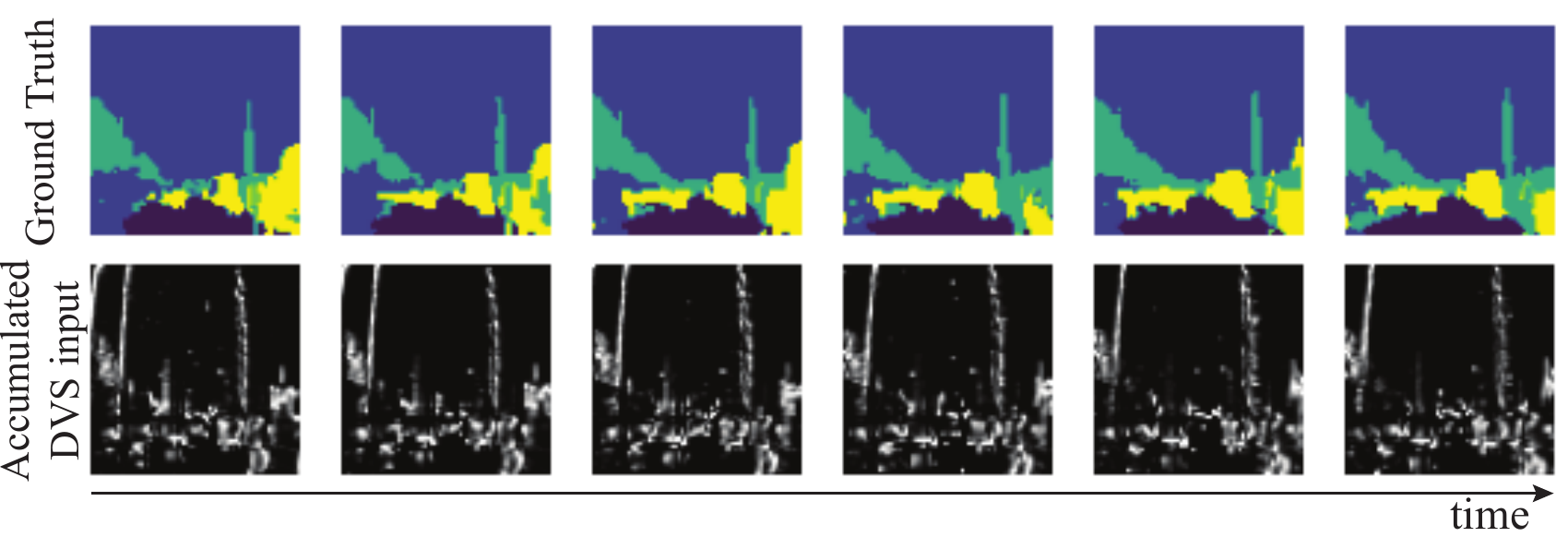} \\
  \vspace{-2mm}
 {\hspace{4mm}(b) DDD17}\\
  \end{center}
  \vspace{-2mm}
  \caption{Examples of input representations. (a) We convert a static image into Poisson spikes. As time goes on, the accumulated spikes represent similar image to original image.
  (b) For the event-based camera input, we accumulate spikes in the pre-defined time window (\eg 10ms) to generate frames.
  } 
   \label{fig:relatedwork:datasetvisualization}
\end{figure}

\begin{figure*}[t!]
  \begin{center}
    \includegraphics[width=0.9\textwidth]{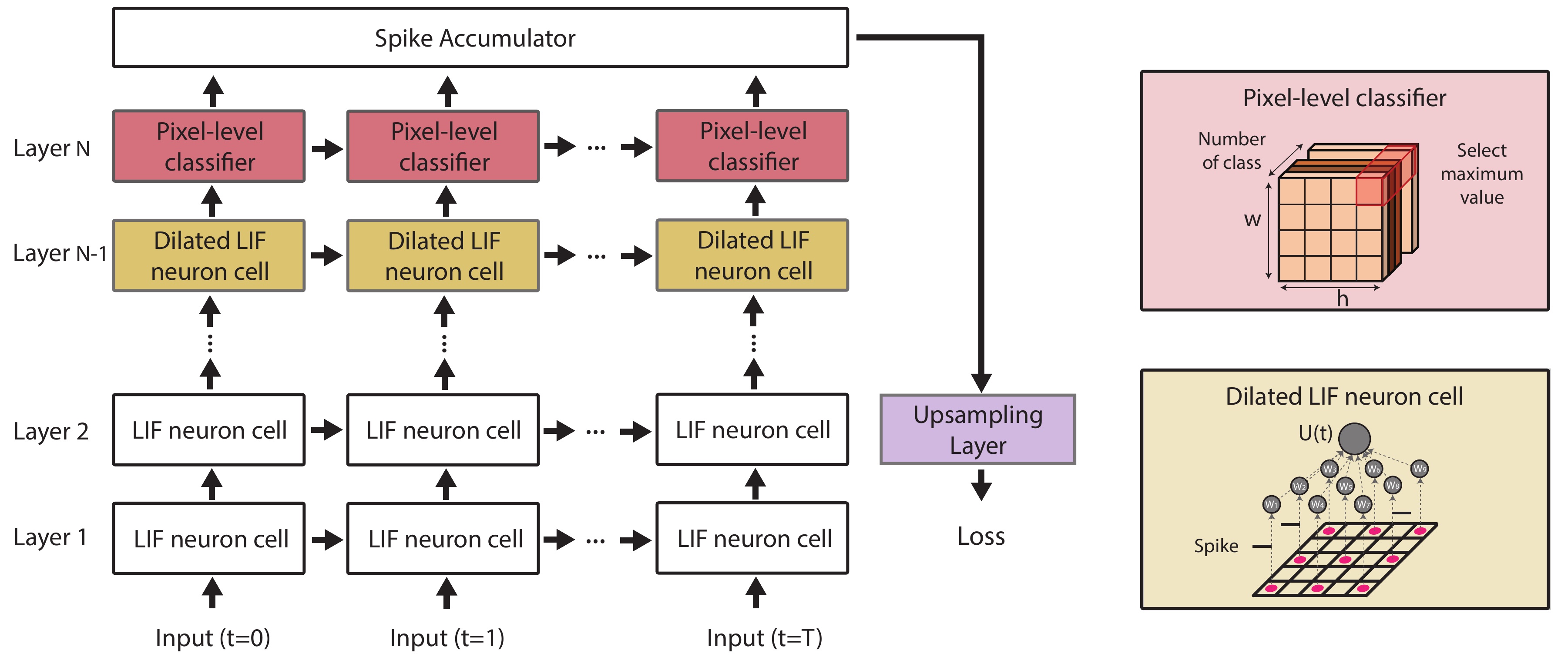}
  \end{center}
  \caption{Spiking-DeepLab: a computational graph unrolled over multiple time-steps. Each neuron cell has a membrane potential consisting of temporal voltage propagation  (\ie horizontal arrow) and  the spikes from previous layer (\ie vertical arrow).
  In order to increase the receptive field in deep layers, we add a dilated LIF neuron cell which can cover a larger spatial area with the same number of parameters.
  For the last layer, we accumulate output spikes across all time-steps and generate a 2-dimensional probabilistic map. Finally, we upsample the final prediction map to the original resolution of the input image and calculate the loss function to update the weights.
  } 
   \label{fig:method:deeplab_architecture}
\end{figure*}


\subsection{Input Representation}
\label{ssec:input_representation}
In this paper, we use two types of input datasets, \ie static images and DVS data.
For training and inference, a static image needs to be converted into spike trains since SNNs process multiple binary spikes. 
There are various spike coding schemes such as rate, temporal, and phase \cite{mostafa2017supervised,kim2018deep}.
Among them, we use rate coding due to its reliable performance across various tasks.
Rate coding provides spikes proportional to the pixel intensity of the given image.
In order to implement this, following previous work \cite{roy2019towards}, we compare each pixel value with a random number ranging between $[I_{min}, I_{max}]$ at every time-step. Here, $I_{min}$ and $I_{max}$ correspond to the minimum and maximum possible pixel intensities.
If the random number is greater than the pixel intensity, the Poisson spike generator outputs a spike with amplitude $1$.
Otherwise, the Poisson spike generator does not yield any spikes. 
We visualize rate coding in Fig. \ref{fig:relatedwork:datasetvisualization}(a). We see that the spikes generated at a given time-step is random. However, as time goes on, the accumulated spikes represent a similar result to the original image.
For a DVS spike stream, we accumulate spikes in a certain time window to generate a frame. Then, the network processes these frames like a video input  (Fig. \ref{fig:relatedwork:datasetvisualization}(b)).

\subsection{Spiking Segmentation Network Architecture}

\subsubsection{Spiking-DeepLab}

Different from image recognition tasks, segmentation networks provide pixel-wise classification with respect to the given 2-dimensional input image.
Therefore, the segmentation architecture needs to maintain a large spatial resolution of the feature maps at the end of the network.
Also, taking a large receptive field helps the networks figure out the relationships between objects in the scene, resulting in high performance.  
To this end, DeepLab \cite{chen2017deeplab} proposed a dilated convolution in order to fulfill these two objectives.
The dilated convolution puts space between the kernel weights.
As a result, the dilated convolution operation increases the size of the receptive field without adding a memory burden.

With the LIF neuron model, for a neuron at ${(i_x, i_y)}$, we can reformulate the neuronal dynamics (Eq. \ref{eq:LIF}) with the dilated convolution operation:
\begin{equation}
    u_{(i_x, i_y)}^t = \lambda u_{(i_x, i_y)}^{t-1} + \sum_{m=1}^K \sum_{n=1}^K  w_{(m,n)} o^t_{(i_x+m\cdot r -1, i_y+n \cdot r -1)}.
    \label{eq:dilated_conv}
\end{equation}
where, $r$ is stride and $K$ is kernel size. If we set $r$ to 1, this equation is the same equation of a normal convolutional layer. If we increase the value of $r$, the receptive fields also increase proportional to $r$.
For example, the area of a local receptive field with $3 \times 3$ kernel is $9$. However, for a dilated convolutional layer with stride $2$, the area of local receptive field increases to $25$, as shown in the right bottom box in Fig. \ref{fig:method:deeplab_architecture}. 
We apply these dilated convolutional layers to the last two convolutional layers in the feature extractor.

Fig. \ref{fig:method:deeplab_architecture} illustrates the computational graph of Spiking-DeepLab.
According to Eq. \ref{eq:dilated_conv}, the membrane potential of each neuron is computed by combining the previous membrane potential (\ie temporal propagation) and the previous layer (\ie spatial propagation).
To preserve the size of the feature map in deep layers, we do not apply average pooling layers in the last feature extraction block (architecture details are shown in Table \ref{table:architecture_detail}).
Also, it is worth mentioning that we use a pixel-wise classifier at the last layer of the network, resulting in a tensor in which the channel size is the number of classes.
We select the class corresponding to the maximum activation at each pixel location.

\subsubsection{Spiking Fully-Convolutional Networks (FCN)}

\begin{figure}[t!]
  \begin{center}
    \includegraphics[width=0.5\textwidth]{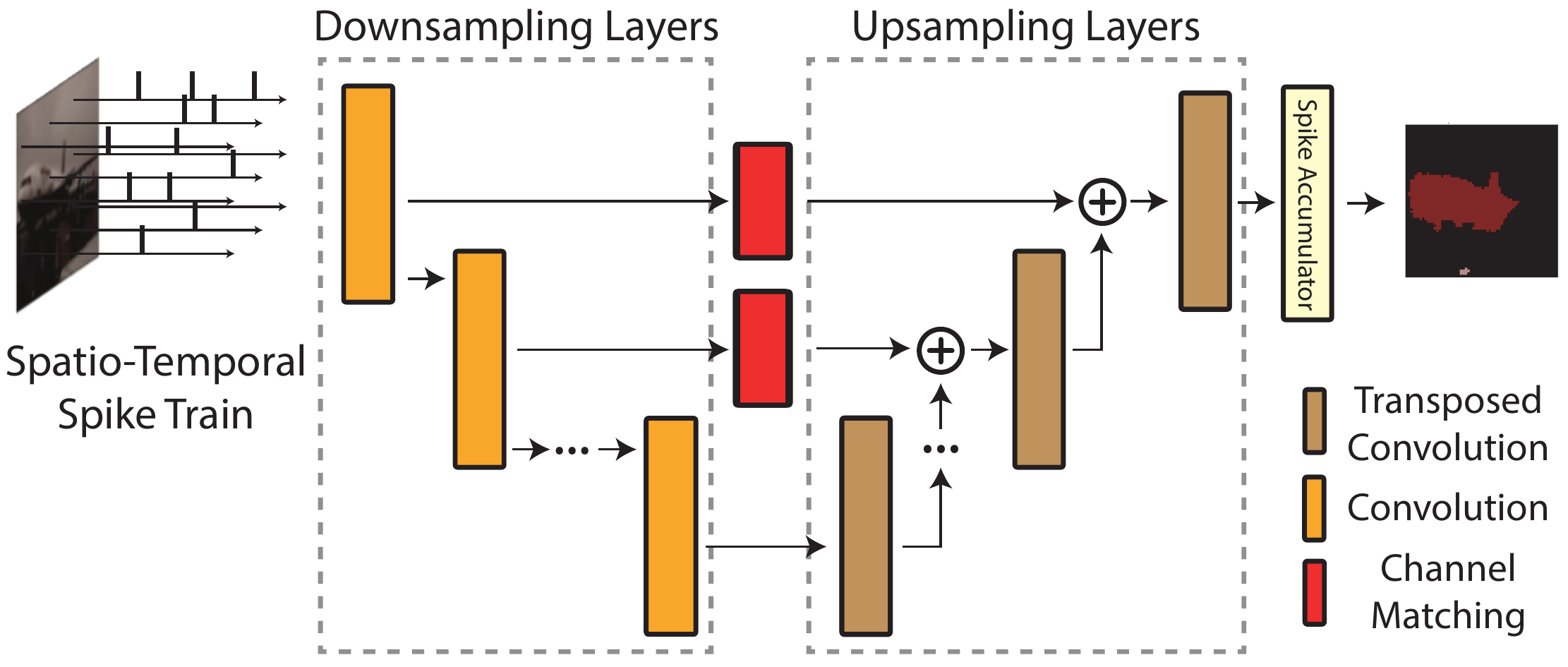}
  \end{center}
  \caption{ An illustration of the FCN architecture. } 
   \label{fig:method:fcn_architecture}
\end{figure}

Another fundamental architecture for segmentation is Fully-Convolutional Networks (FCN). Different from DeepLab where a high resolution feature map is maintained, FCN consists of a downsampling network and an upsampling network.
The downsampling network discovers the semantic representation of the given input images.
This is similar to image recognition networks which consist of multiple convolution and pooling layers.
The upsampling network recovers the resolution of the small feature map by upsampling it into the original image resolution using transposed convolutional layers.
The transposed convolutional layer increases the resolution of the input feature map to a desired output feature map size with learn-able kernel parameters. These parameters are also trained with gradient-based learning.
While increasing the resolution, following the original FCN paper \cite{long2015fully}, we add intermediate features from the downsampling network to features from the upsampling network. 
Here, we use convolutional layers to match the number of channels between features in downsampling and upsampling.
Note that all layers consist of LIF neurons, and thus features are processed in a temporal manner.
In addition, like the DeepLab architecture, we use the spike accumulator at the end of the network.
We visualize the architecture (not the computational graph) of FCN in Fig. \ref{fig:method:fcn_architecture}.
The computational graph of Spiking-FCN can be represented with a similar structure as Spiking-DeepLab.

\subsection{Overall Optimization}

For both architectures, the network provides a 2-dimensional probabilistic map at the output layer.
This probabilistic map is based on the stacked spike voltage from the spike accumulator.
For every pixel location $(p, q)$, with the given ground truth label $y_i$, we compute the cross-entropy loss as: 
\begin{equation}
    {L} = - \frac{1}{N} \sum_{p, q} \sum_{i} y_{i}^{(p,q)} log(\frac{e^{h_i^{(p,q)}}}{\sum_{k=1}^{C}e^{h_k^{(p,q)}}}).
    \label{eq:celoss}
\end{equation}
Here, $N$ and $C$ represent a normalization factor and  the number of classes, respectively. Also, $h_k$ stands for the membrane potential of the neurons in the last layer. We calculate backward gradients based on this spatial cross-entropy loss.

Backward gradients are computed based on surrogate back-propagation through time (BPTT) \cite{neftci2019surrogate,wu2018spatio} in which gradients are accumulated over all time-steps across all layers.
For a hidden layer, the gradient of output spikes with respect to a membrane potential (\ie $\frac{\partial o_i^t}{\partial u_i^t}$) is not differentiable because of the firing behavior of the LIF neuron (Fig. \ref{fig:relatedwork:neuron_dynamic}).
To enable backpropagation through multiple layers, we approximate the backward gradient function to a piece-wise linear function:
\vspace{-1mm}
\begin{equation}
    \frac{\partial o_i^t}{\partial u_i^t} =  \max \{0, 1-  \ | \frac{u_i^t - \theta}{\theta} \ | \},
    \label{eq:approximated_gradient}
\vspace{-1mm}
\end{equation}
For the last layer (\ie classifier), weight parameters can be updated by conventional gradient backpropagation since we accumulate the spikes and conduct a single forward step in the last layer.
Overall, weight parameters are updated based on the accumulated gradients at each layer.
The accumulated gradients of loss $L$ with respect to weights $W_l$ at layer $l$ can be calculated as:  
\begin{equation}
      \frac{\partial L}{\partial W_l} =
\begin{cases}
 \sum_{t}(\frac{\partial L}{\partial O_l^t}\frac{\partial O_l^t}{\partial U_l^t} + \frac{\partial L}{\partial U_l^{t+1}}  \frac{\partial U_l^{t+1}}{\partial U_l^{t}})
 \frac{\partial U_l^t}{\partial W_l},  & \text{ $l$: hidden layer} \\
    \frac{\partial L}{\partial h}\frac{\partial h}{\partial W_{fc}}.
    & \text{$l$: output layer} 
\end{cases}
\label{eq:delta_W}
\end{equation}
Here, $O_l$ and $U_l$ stand for the output spike matrix and membrane potential matrix at layer $l$, respectively. We refer to \cite{wu2018spatio} for more spatio-temporal backpropagation details.
Algorithm 1 shows the overall optimization process of spiking segmentation networks with surrogate gradient backpropagation.

\begin{algorithm}[t]\small
        \caption{Directly training a segmentation network with surrogate gradient backpropagation}
       \textbf{Input}: spike time-step ($T$); mini batch ($X$);  labels ($Y$)\\
      \textbf{Output}:  spiking segmentation network
      \begin{algorithmic}[1]
        %
        \For{$i \gets 1$ to $max\_iter$}
        \State {fetch a mini batch $X$}
        \For{$t \gets 1$ to $T$}  
            \State{O $\leftarrow$ PoissonGenerator(X) (or DVS input X)}
            \For{$l \gets 1$ to $L-1$} 
                \State{$(O^t_{l}, U_l^{t}) \leftarrow (\lambda, U_l^{t-1}, (W_{l}, O^{t}_{l-1}))$}
                \Comment{Eq. (\ref{eq:dilated_conv})}
            \EndFor
            \State{$U_L^{t} \hspace{-1mm} \leftarrow  \hspace{-1mm} ( U_L^{t-1},  (W_{l}, O^{t}_{L-1}))$ } \Comment{Final layer}
        \EndFor
        \State{$L \leftarrow (U_L^T, Y)$} 
        \State{Do back-propagation and weight update}
    \EndFor
    
      \end{algorithmic}
          \label{algorithm: overall}
\end{algorithm}

\section{Experiments}
    \addtolength{\tabcolsep}{2.5pt}
\begin{table*}[t!]
\centering
\caption{SNN architectures of DeepLab and FCN.}
\resizebox{0.7\textwidth}{!}{
\begin{tabular}{|c|c|c|}
\hline
Module & Spiking-DeepLab & Spiking-FCN \\ 
\hline
& \makecell{Conv [64] \\
         Conv [64]  \\
          AvgPool [stride 2]}
& \makecell{Conv [64] \\
         Conv [64]  \\
          AvgPool [stride 2]}
\\ \cline{2-3}
\makecell{Downsampling \\ Layers} & \makecell{Conv [128] \\
         Conv [128]  \\
          AvgPool [stride 2]}
& \makecell{Conv [128] \\
         Conv [128]  \\
          AvgPool [stride 2]}
\\ \cline{2-3}
& \makecell{Conv [256] \\
         Dilated Conv [256]  \\
         Dilated Conv [256]}
& \makecell{Conv [256] \\
         Conv [256]  \\
         Conv [256]  \\
          AvgPool [stride 2]}
\\ \hline
\makecell{Intermediate \\ Layers}
& \makecell{Conv [1024] \\
         Conv [1024]  \\
         Conv [1024, Number of classes]}
& \makecell{Conv [1024] \\
         Conv [1024]}
\\ \hline
\makecell{Upsampling \\ Layers}
& \makecell{Bilinear interpolation}
& \makecell{Conv [1024, Number of classes] \\
         TransposeConv [Number of classes]  \\
         SkipConv [256, Number of classes] \\
         TransposeConv [Number of classes] \\
         SkipConv [128, Number of classes] \\
         TransposeConv [Number of classes] \\
         SkipConv [64, Number of classes]}
\\ \hline
\end{tabular}}
\label{table:architecture_detail}
\end{table*}

\subsection{Experimental Setup}

\subsubsection{Dataset}

We evaluate our methods on two semantic segmentation datasets: PASCAL VOC2012 and DDD17.

\textbf{PASCAL VOC2012} \cite{everingham2010pascal} contains static images taken with a conventional camera classified with 20 foreground object classes and one background class. We use an augmented dataset \cite{hariharan2011semantic} consisting of a training split of 10,582 images and a validation split of 1,449 images. We re-scale the various original image resolutions to 64$\times$64 pixels.
For SNN models trained and evaluated on PASCAL VOC2012, we use a rate coding technique (discussed in Section \ref{ssec:input_representation}).
Note, we use \textit{PASCAL VOC2012} and \textit{VOC2012} interchangeably in the remainder of the paper.

\textbf{DDD17} \cite{binas2017ddd17} contains 40 different driving sequences of event data captured by a DVS camera. While the dataset provides both grayscale images and event data, it does not provide semantic segmentation labels. Therefore, we use the segmentation labels provided in \cite{alonso2019ev}, which consists of 20 different sequence intervals in 6 of the original DDD17 sequences.
From these sequence intervals, we use a training split consisting of 15,950 frames and a testing split consisting of 3,890 frames with 6 classes. We use a two-channel event representation of accumulated positive and negative events (integrated for a time interval of 50ms) from the DVS dataset proposed in \cite{alonso2019ev}, and we re-scale the image resolution of 346$\times$200 pixels to 64$\times$64 pixels.
In order to train SNNs with DDD17, we do not apply the Poisson coding technique since DDD17 contains temporally-related sequential samples.
Therefore, we construct a batch with multiple sequences of successive video frames to leverage temporal information.
For ANNs, we give each frames to networks, and average all embedding features (after the feature extractor) in temporal axis.

\begin{table}[t]
    \addtolength{\tabcolsep}{2.5pt}
    \centering
     \caption{Mean IoU (\%) of ANNs, Spiking-FCN, and Spiking-DeepLab on PASCAL VOC2012.}
    \resizebox{0.48\textwidth}{!}{%
    \begin{tabular}{lcc}
        \toprule
        Method   &  Time-steps  &   MIoU(\%) \\
        \midrule
           DeepLab \cite{chen2017deeplab}    &  -  &  {32.3} \\
          FCN \cite{long2015fully} &  -  &  30.9  \\
         Spiking-DeepLab (ours)  &  20  &  {22.3}  \\
         Spiking-FCN (ours)    &  20  &  9.9  \\
        \bottomrule
        \\
    \end{tabular}}
    \label{table:exp:performance_comparison_VOC2012}
\end{table}

\begin{table}[t]
    \addtolength{\tabcolsep}{2.5pt}
    \centering
     \caption{Mean IoU (\%) of ANNs, Spiking-FCN, and Spiking-DeepLab on DDD17.}
    \resizebox{0.48\textwidth}{!}{%
    \begin{tabular}{lcc}
        \toprule
        Method   &  Time-steps  &   MIoU(\%) \\
        \midrule
           DeepLab \cite{chen2017deeplab}    &  -  &  33.8 \\
          FCN \cite{long2015fully} &  -  &   40.2  \\
         Spiking-DeepLab (ours)  &  20  &  33.7 \\
         Spiking-FCN (ours)    &  20  & 34.2  \\
        \bottomrule
        \\
    \end{tabular}}
    \label{table:exp:performance_comparison_DDD17}
\end{table}

\subsubsection{Parameters}

In our experiments, we use two different types of SNNs as shown in Table \ref{table:architecture_detail}.
For both networks, the backbone network consists of 7 layers since it is difficult to increase depth with SNNs due to the real vs. surrogate gradient discrepancy.
Note that the main advantage of SNNs is huge energy efficiency compared to ANN, which is desirable for artificial Intelligence systems in edge devices.
A Spiking-DeepLab model includes two dilated convolutional layers in the backbone network followed by a 3-layer classifier.
For Spiking-DeepLab, we use bilinear interpolation to upsample the output prediction to a segmentation map.
Our Spiking-FCN model is similar to our Spiking-DeepLab model but does not use dilated convolutions and uses transposed convolutional layers for upsampling.
\textit{SkipConv} denotes the skip connection with channel reduction (red block in Fig. \ref{fig:method:fcn_architecture}).
We configure the same ANN architectures for comparison. In the ANNs, we use batch normalization \cite{ioffe2015batch} at each convolutional layer. For SNNs, we use the temporal Batch Normalization Through Time (BNTT) technique \cite{kim2020revisiting}.

We train both networks using the same hyperparameters.
We use an Adam optimizer with learning rate 3e-3.
We use a batch size of 16.
Also, we use step-wise learning rate scheduling with a decay factor of 10 at 50\% of the total number of epochs. Here, we set the total number of epochs to 60.
We set 20 time-steps, set a leak factor of 0.99, and use a membrane threshold of 1.0.


\begin{figure*}[t!]
  \begin{center}
    \includegraphics[width=0.8\textwidth]{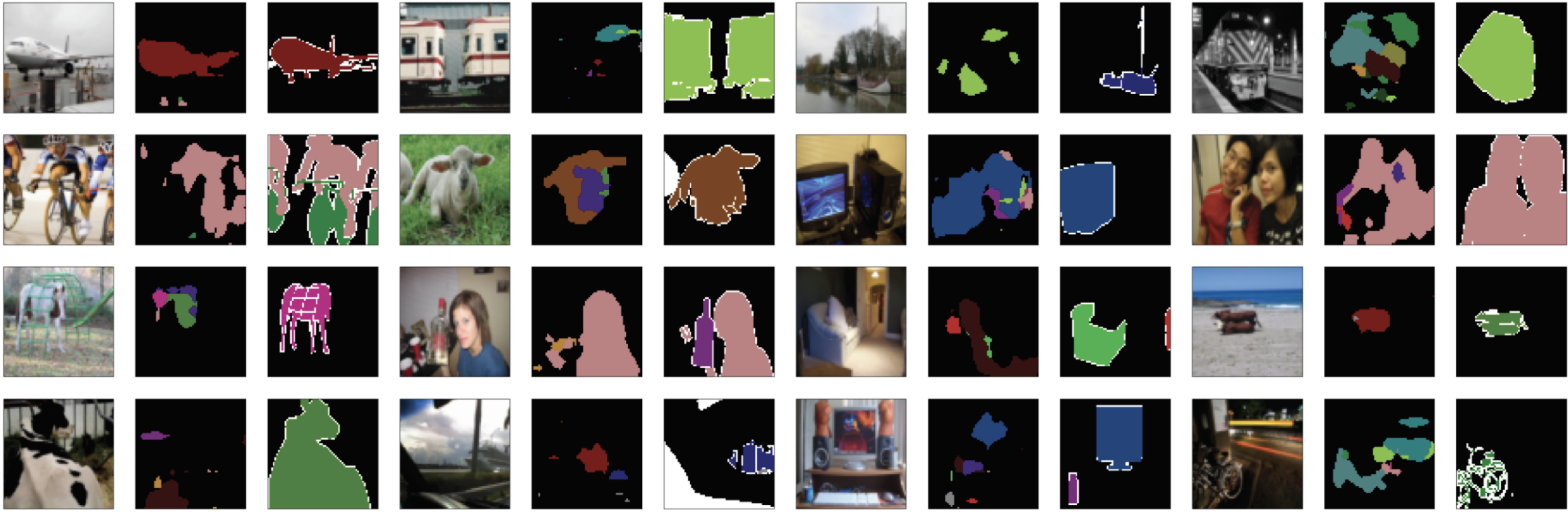}\\
    {\hspace{4mm}(a) Spiking-DeepLab} \\ 
    \vspace{2mm}
     \includegraphics[width=0.8\textwidth]{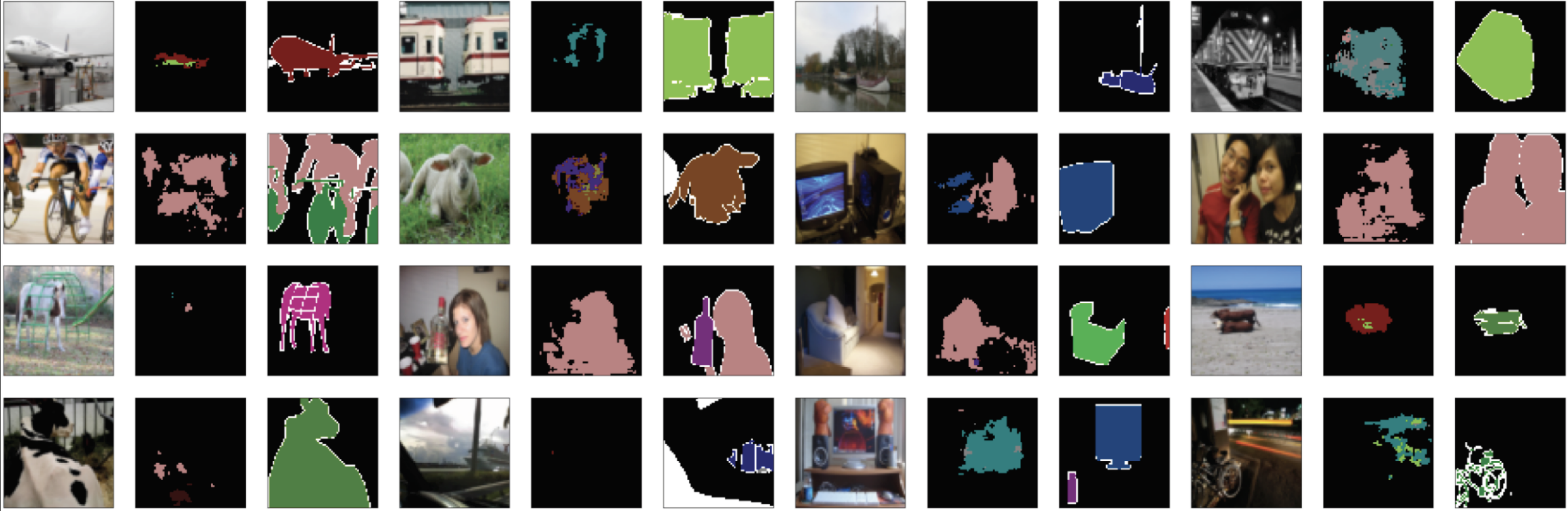} \\
     {\hspace{4mm}(b) Spiking-FCN} \\ 
  \end{center}
  \caption{ Qualitative results of Spiking-DeepLab and Spiking-FCN on the PASCAL VOC2012 validation set. 
    We visualize the \textit{image-prediction-groundtruth} triplet for each sample.}
   \label{fig:exp:visualization_voc}
\end{figure*}
\begin{figure*}[t!]
  \begin{center}
    \includegraphics[width=0.8\textwidth]{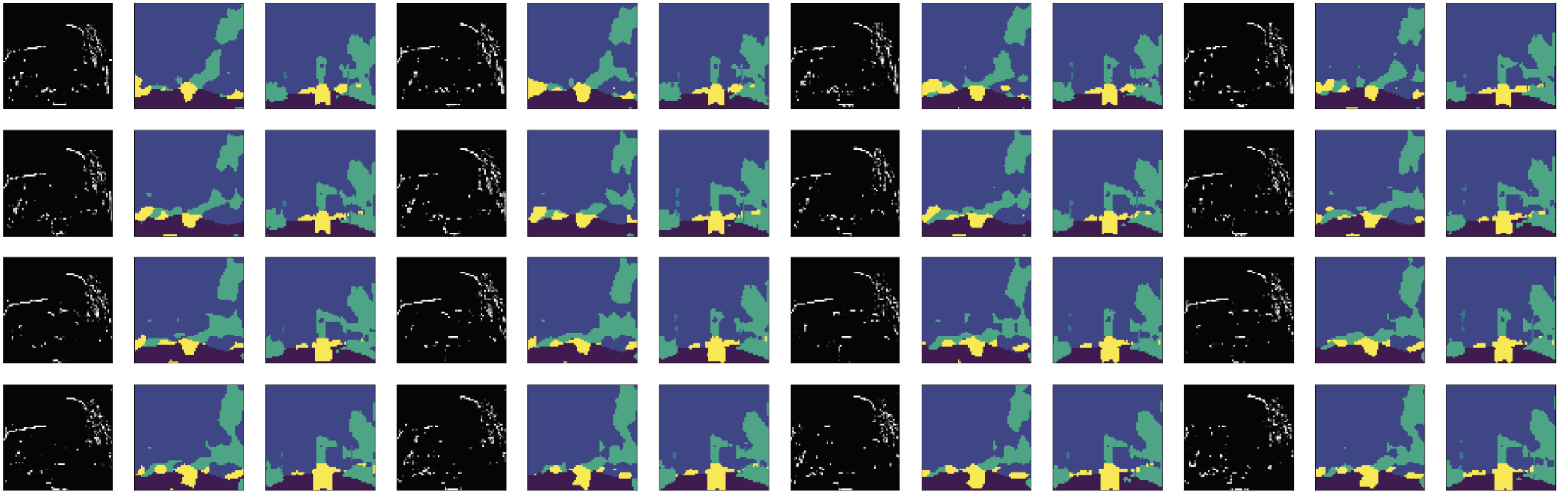}\\
    {\hspace{4mm}(a) Spiking-DeepLab} \\ 
    \vspace{2mm}
     \includegraphics[width=0.8\textwidth]{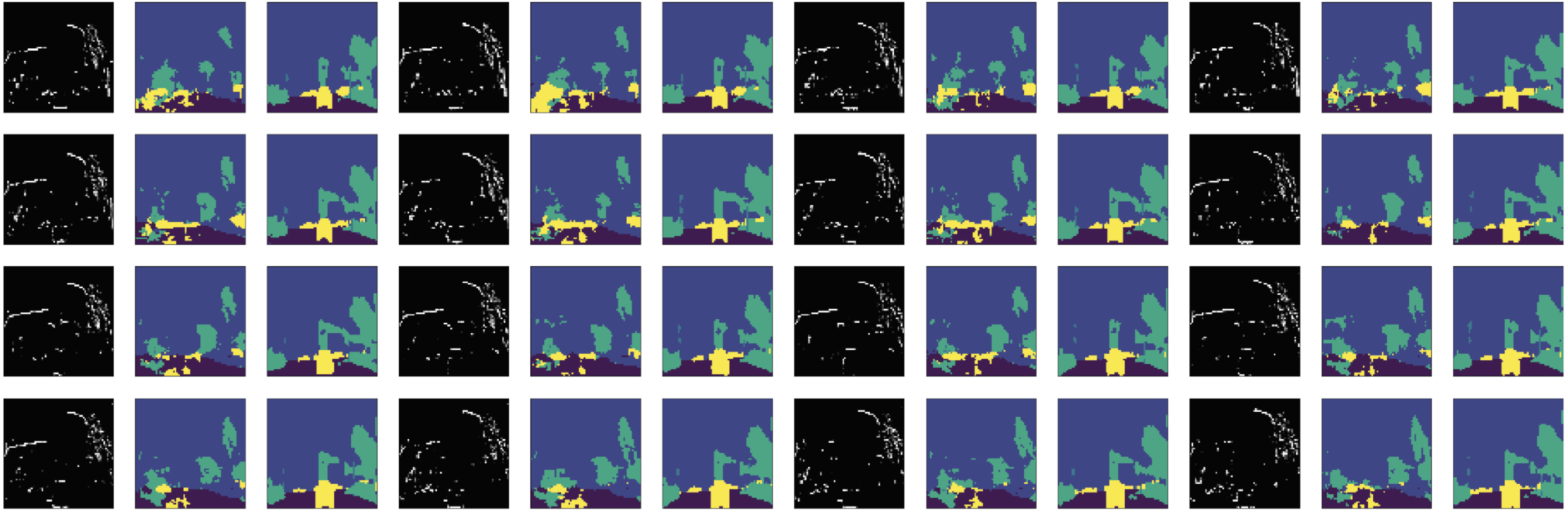} \\
     {\hspace{4mm}(b) Spiking-FCN} \\ 
  \end{center}
  \caption{ Qualitative results of Spiking-DeepLab and Spiking-FCN on the DDD17 test set.
  We visualize the \textit{image-prediction-groundtruth} triplet for each sample.  } 
   \label{fig:exp:visualization_ddd17}
\end{figure*}

\begin{figure}[t]
\begin{center}
\def\arraystretch{0.5}
\begin{tabular}{@{}c@{\hskip 0.05\linewidth}c@{}c}
\includegraphics[width=0.9\linewidth]{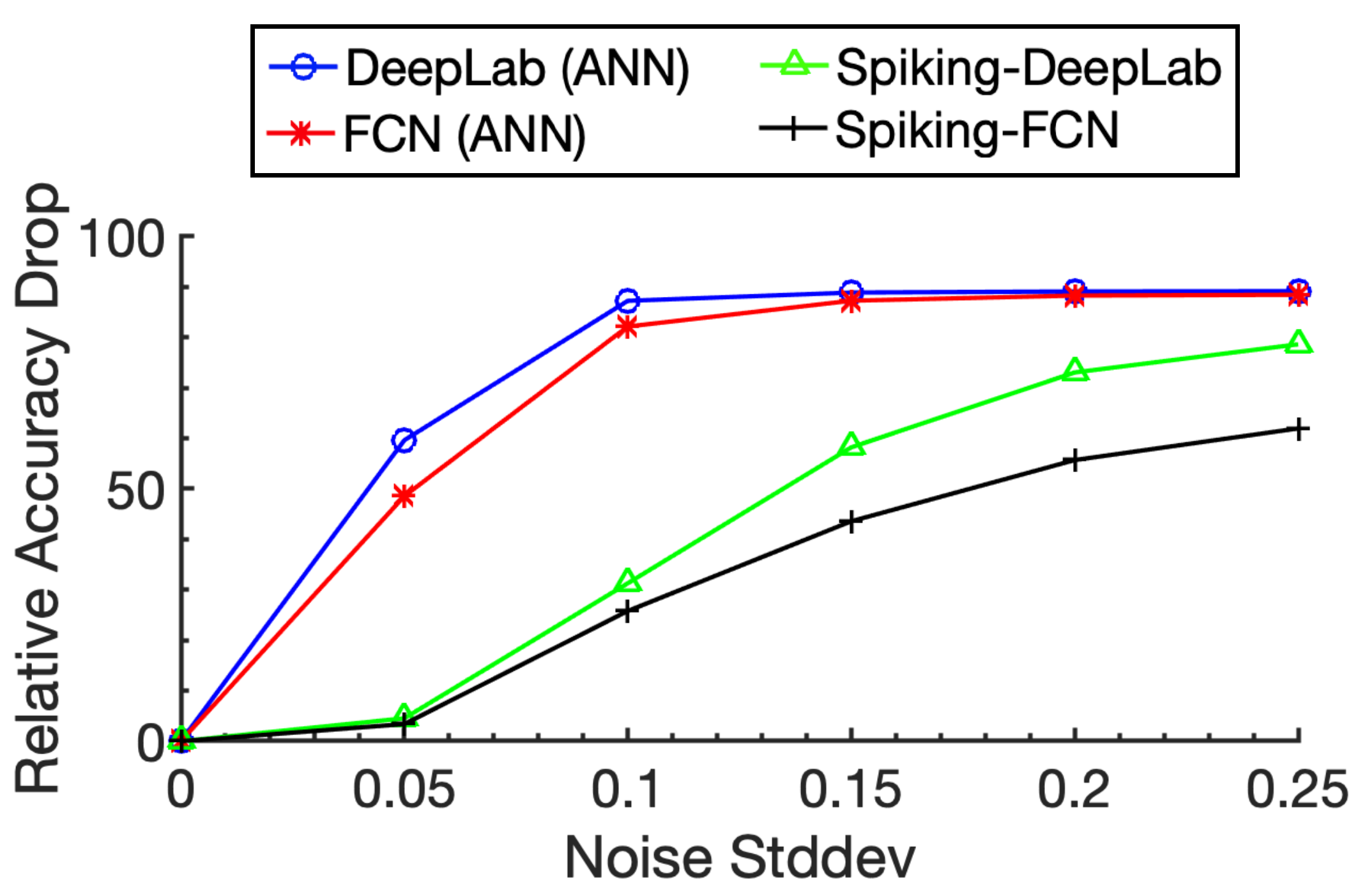} \\
{\hspace{4mm}(a) PASCAL VOC2012} \\ 
\vspace{1mm}
\includegraphics[width=0.9\linewidth]{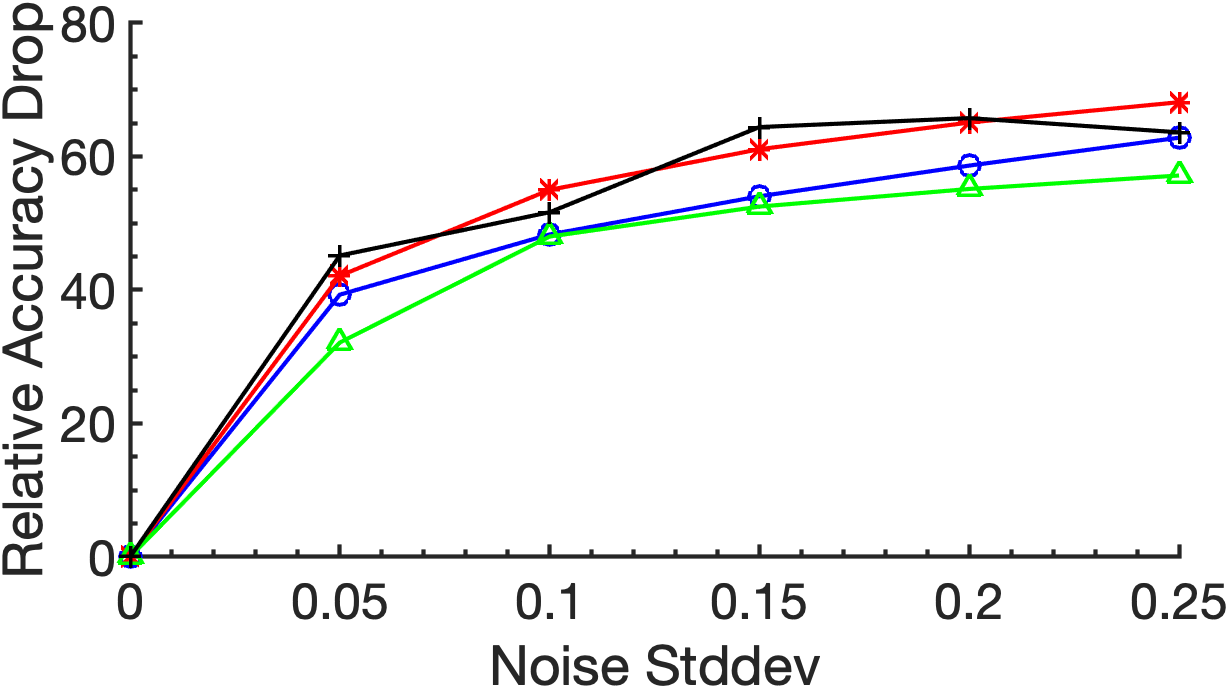} \\
 {\hspace{4mm}(b) DDD17}\\
\end{tabular}
\end{center}
\vspace{0mm}
\caption{Comparison of changes in Mean IoU with respect to varying standard deviations of Gaussian noise.
We use relative accuracy drop ($\frac{CleanMIoU - NoiseMIoU}{CleanMIoU} \times 100$) where $Clean MIoU$ and $NoiseMIoU$ denote MIoU with clean and noise samples, respectively.
The lower the relative accuracy drop is, the more robust the model is at that noise level.}
\label{fig:noisy_test}
\end{figure}

\subsection{Performance Comparison}

On public datasets, we compare the proposed Spiking-DeepLab and Spiking-FCN methods with the reference ANN methods \cite{he2016deep,long2015fully}.
In Table \ref{table:exp:performance_comparison_VOC2012}, for our BNTT-surrogate gradient-based approaches, the Spiking-FCN architecture shows less performance than the Spiking-DeepLab counterpart on VOC2012. This is due to the FCN architecture being deeper than the DeepLab architecture. Also earlier works have shown that training SNNs with deeper architectures is more challenging than with shallower architectures \cite{panda2020toward,kim2020revisiting}.
In contrast, the Spiking-FCN architecture showed similar performance to the Spiking-DeepLab on DDD17, as shown in Table \ref{table:exp:performance_comparison_DDD17}. This is because of the low level of difficulty for segmentation on the DDD17 dataset which is due to lower number of classes and more consistent, structured data (\ie cars are always located in the middle of images, tress are located in the side of images).
Overall, for both datasets, ANN references have a higher performance than SNNs since they are based on a well-established training method. 
It is worth mentioning that our purpose is to expand the application of SNNs and show the feasibility of various training methods. In the following sections, we show SNNs have the advantages of robustness and energy-efficiency.
We also visualize the segmentation results on both datasets in Fig. \ref{fig:exp:visualization_voc} and Fig. \ref{fig:exp:visualization_ddd17}.

\subsection{Analysis on Robustness}

In real-world applications such as a self-driving vehicle, input signals are likely to be susceptible to noise.
To investigate the effect of SNNs on robustness for segmentation, we evaluate the relative accuracy drop ($\frac{CleanMIoU - NoiseMIoU}{CleanMIoU} \times 100$) in MIoU across varying levels of noise. 
Here, $CleanMIoU$ and $NoiseMIoU$ denote model MIoU when given clean samples and noisy samples, respectively.
We add Gaussian noise (0, $\sigma$) to our inputs to generate noisy inputs.
From Fig. \ref{fig:noisy_test}(a), we observe that our SNNs (\ie Spiking-DeepLab and Spiking-FCN) are more robust than ANNs for segmentation on VOC2012.
This is because the Poisson spike generator converts a static pixel value into multiple temporal spikes with random distribution.
From Fig. \ref{fig:noisy_test}(b), we observe that SNNs show a similar robustness with ANNs on DDD17.
Thus, SNNs trained on DDD17 (\ie DVS event stream data) are less robust than SNNs trained on VOC2012 (\ie static image).
The main two factors contributing to this decrease in robustness are the following:
(i) As aforementioned, the SNNs trained and evaluated on DDD17 do not use the Poisson coding technique which greatly affects robustness. 
(ii) The predictions of SNNs on DDD17 are based on multiple sequential frames whereas ANNs are only fed one input for each prediction.
Therefore, more noise is accumulated across multiple frames for SNNs.
However, despite the reduced robustness for segmentation on DVS data, segmentation networks with spiking neurons offer improved robustness for segmentation using traditional cameras.

\subsection{Analysis on Energy-efficiency}

In addition to robustness, SNNs are well known for high energy-efficiency compared to ANNs.
In order to verify this, we compute the approximated energy consumption of ANNs and SNNs. It is worth mentioning that we neglect memory and any peripheral circuit energy and consider the energy needed only for Multiply and Accumulate (MAC) operations.

\begin{figure}[t]
\begin{center}
\def\arraystretch{0.5}
\begin{tabular}{@{}c@{\hskip 0.05\linewidth}c@{}c}
\includegraphics[width=0.9\linewidth]{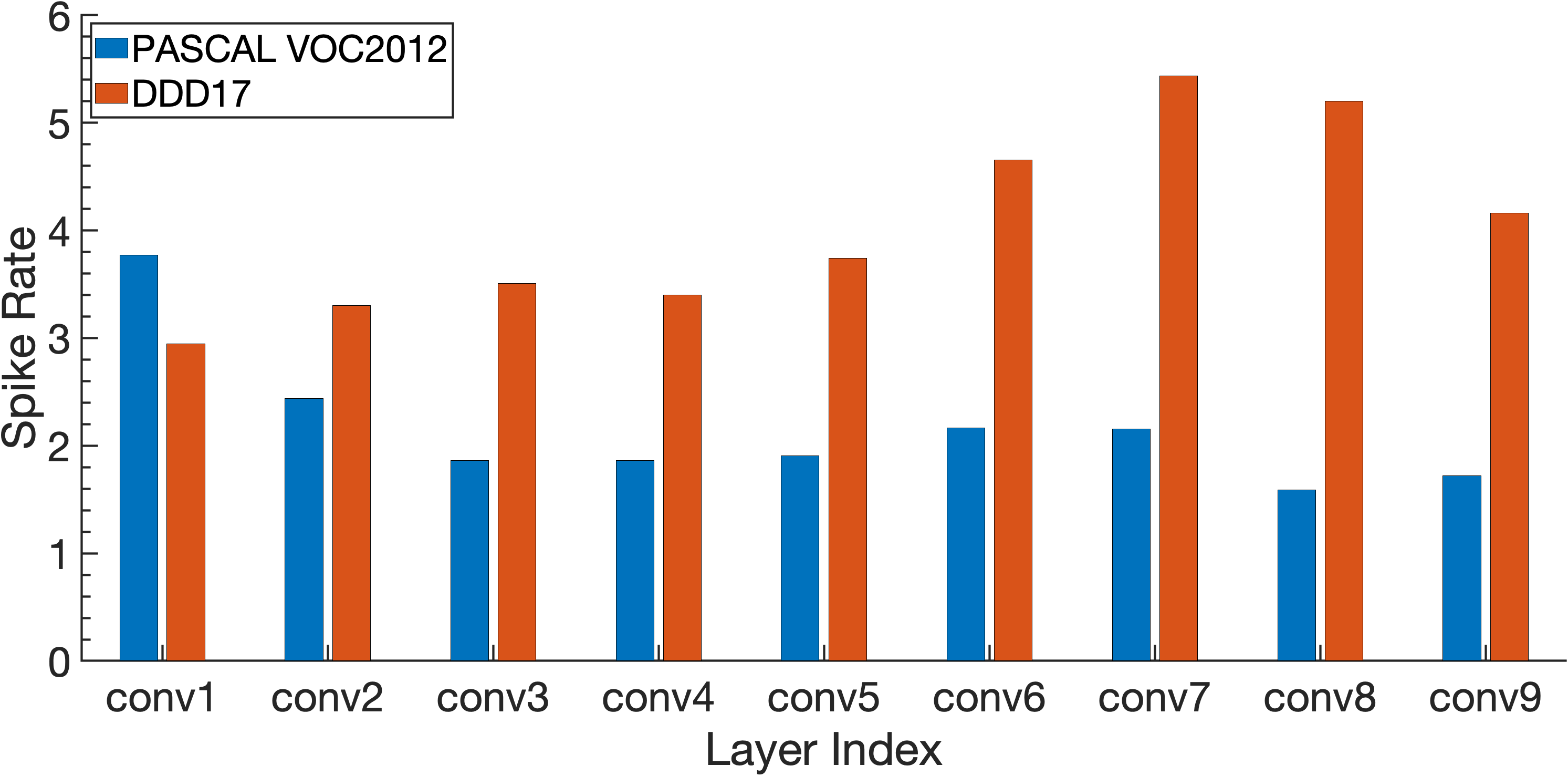} \\
{\hspace{4mm}(a) Spiking-DeepLab} \\ 
\includegraphics[width=0.9\linewidth]{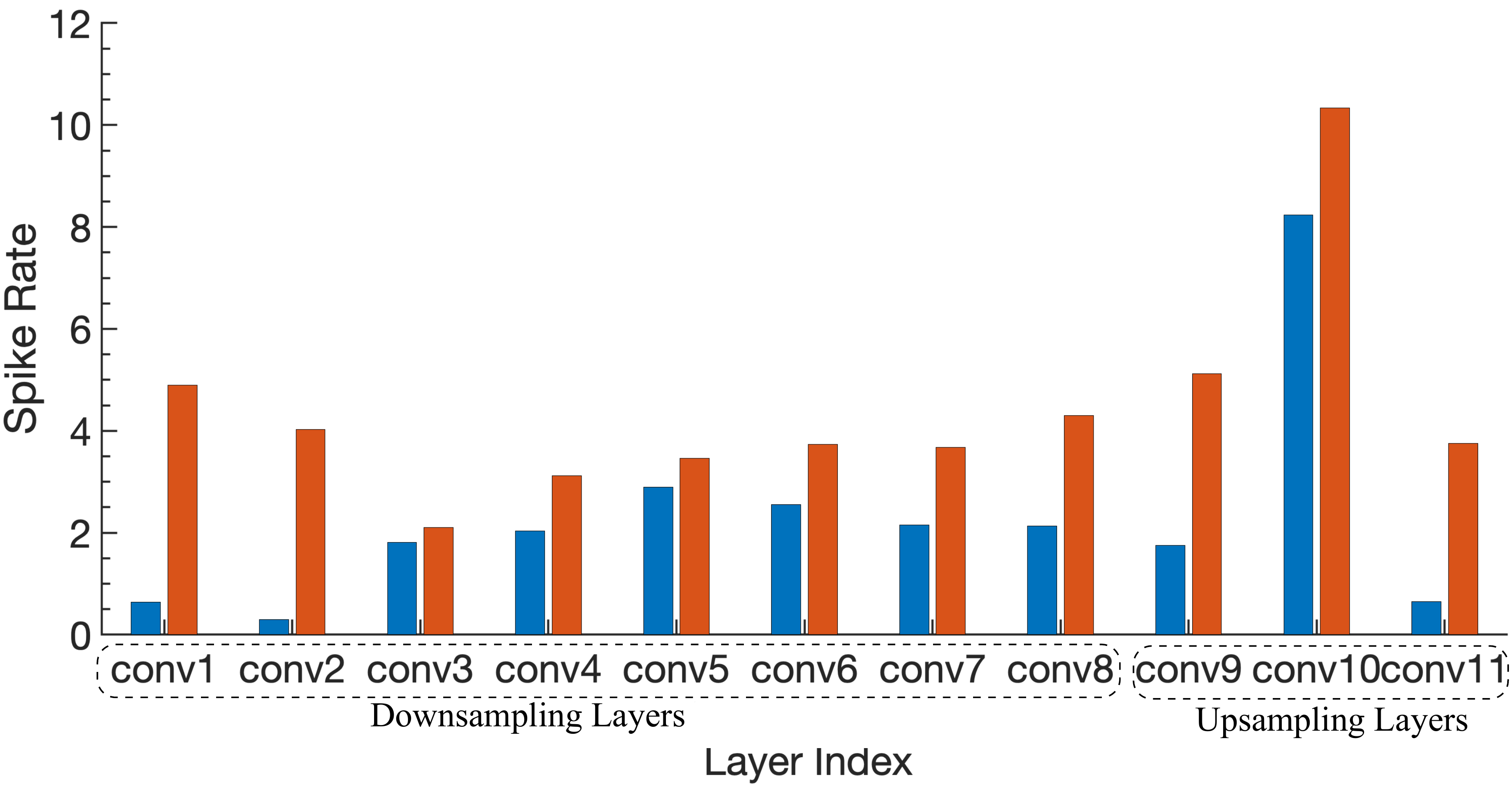} \\
 {\hspace{4mm}(b) Spiking-FCN}\\
\end{tabular}
\end{center}
\caption{Spike rate across all layers in (a) Spiking-DeepLab and (b) Spiking-FCN. We calculate the spike rate for both architectures on PASCAL VOC2012 and DDD17.
}
\label{fig:exp:spikerate}
\end{figure}

Since MAC operations are based on layer-wise spiking rates, we measure the layer-wise spiking rates of Spiking-DeepLab (Fig. \ref{fig:exp:spikerate}(a)) and Spiking-FCN (Fig. \ref{fig:exp:spikerate}(b)) on two benchmarks (\ie PASCAL VOC2012 and DDD17).  
Specifically, we calculate the spike rate $R_s(l)$ of each layer $l$, which can be defined as the total number of spikes at layer $l$ over total time-steps $T$ divided by the number of neurons in layer $l$:

\begin{equation}
    R_s(l) = \frac{\textup{$\#$spikes of layer $l$ over all time-steps}}{\textup{ $\#$neurons of layer $l$}}.
    \label{eq:spking_rate}
\end{equation}

Interestingly, in Fig. \ref{fig:exp:spikerate}(b), we observe high spike rates in the upsampling layers.
This is because of skip connections in Spiking-FCN which inject a number of spikes during forward propagation.

More precisely, following the previous works \cite{park2020t2fsnn,lee2016training,rathi2020diet}, we compute the energy consumption for SNNs by calculating the total number of floating point operations (FLOPs).
Also, to compare ANNs and SNNs quantitatively, we calculate the energy based on standard CMOS technology \cite{horowitz20141} as shown in Table \ref{table: cmos_tech}.
Conventional ANNs require one FP addition and one FP multiplication to conduct the same MAC operation \cite{sze2017efficient}.
On the other hand, as the computation of SNNs are event-driven with binary spike processing, the MAC operation reduces to just a floating point (FP) addition. 
For a layer $l$ in ANNs, we can calculate FLOPs as:
\begin{equation}
FLOPs_{ANN}(l) = \left\{\begin{matrix} &  \hspace{-5mm} k^2 \times O^2 \times C_{in} \times C_{out},  \hspace{0.5mm}  \textup{if $l =$ Conv},
\\ 
&\hspace{-3mm}C_{in} \times C_{out}, \hspace{17mm} \textup{if $l =$ Linear}.
\end{matrix}\right.
\end{equation}
Here, $k$ is kernel size, $O$ is output feature map size, and 
$C_{in}$ and $C_{out}$ are input and output channels, respectively.
For SNNs, neurons consume energy whenever the neurons are activated. Thus, we multiply the spiking rate $R_s(l)$ (Eq. \ref{eq:spking_rate}) with FLOPs as: 
\begin{equation}
    FLOPs_{SNN}(l) = {FLOPs}_{ANN}(l) \times R_s(l).
\end{equation}
Finally, the total inference energy of ANNs ($E_{ANN}$) and SNNs ($E_{SNN}$) across all layers can be obtained using the energy values ($E_{MAC}$, $E_{AC}$) from Table \ref{table: cmos_tech}.
\begin{equation}
    E_{ANN} = \sum_{l} FLOPs_{ANN}(l)  \times E_{MAC}.
\end{equation}
\begin{equation}
    E_{SNN} = \sum_{l} FLOPs_{SNN}(l) \times E_{AC}.
\end{equation}

We compare the energy efficiency between ANNs and SNNs on two benchmarks in Table \ref{table: energy_consumption}.
As expected, SNNs show higher energy-efficiency (\ie $E_{ANN}/E_{method}$) compared to ANNs. Moreover, we observe that SNNs bring more energy-efficiency on the static VOC2012 dataset since it generates less numbers of spikes across all layers (Fig. \ref{fig:exp:spikerate}).
Note that, ANNs have the same energy consumption regardless of datasets whereas SNNs have different energy consumption depending on the number of spikes.
It is worth mentioning that SNNs can obtain much higher energy-efficiency on neuromorphic hardware platform \cite{akopyan2015truenorth,furber2014spinnaker}.

\begin{table}[t]
\addtolength{\tabcolsep}{1.5pt}
\centering
\caption{Energy table for 45nm CMOS process.}
\label{table:energy_efficiency}
\resizebox{0.45\textwidth}{!}
{
\begin{tabular}{lc}
\toprule
Operation   & Energy(pJ)  \\
\midrule
    32bit FP MULT $(E_{MULT})$   & 3.7  \\
    32bit FP ADD  $(E_{ADD})$  & 0.9  \\
    32bit FP MAC $(E_{MAC})$   & 4.6 (= $E_{MULT}$ + $E_{ADD}$)  \\
    32bit FP AC  $(E_{AC})$  & 0.9  \\
\bottomrule
\end{tabular}%
}
\label{table: cmos_tech}
\end{table}

\begin{table}[t]
\addtolength{\tabcolsep}{1.5pt}
\centering
\caption{Energy efficiency comparison between segmentation networks.}
\resizebox{0.47\textwidth}{!}{%
\begin{tabular}{lcc}
\toprule
Method &  Dataset & $E_{ANN}/E_{method}$ \\
\midrule
    DeepLab (ANN) \cite{he2016deep}    & - &  1$\times$ (reference) \\
    Spiking-DeepLab  & VOC2012  &  2.75$\times$ \\
    Spiking-DeepLab  & DDD17   &  1.15$\times$ \\
    \midrule
    FCN (ANN) \cite{long2015fully}    & - &  1$\times$ (reference) \\
    Spiking-FCN  & VOC2012  &  2.37$\times$ \\
    Spiking-FCN  & DDD17   &  1.04$\times$ \\
\bottomrule
\end{tabular}%
}
\label{table: energy_consumption}
\end{table}

\section{Conclusion}

In this paper, for the first time, we perform a comprehensive study on the feasibility of training SNNs on the semantic segmentation task.
We construct two representative spiking segmentation networks, \ie Spiking-DeepLab and Spiking-FCN.
We show ANN-SNN conversion requires a large number of time-steps (more than one thousand) and shows inferior performance due to the complexity of the segmentation task.
This is an important observation since most state-of-the-art SNN training techniques for image recognition are based on ANN-SNN conversion.
Instead, we leverage approximated gradient-based training with a temporal batch normalization technique (\ie BNTT), resulting in reasonable segmentation results with a small number of time-steps.
We experimentally evaluate the performance, robustness, and estimated energy on two different semantic segmentation scenarios: the static PASCAL VOC2012 dataset and the DDD17 dataset consisting of event stream spikes.
The proposed segmentation architectures can bring a more than $2\times$ energy efficiency on the static image dataset.
As future work,  we plan to develop an advanced spike-based learning algorithm to fill the performance gap between ANNs and SNNs on semantic segmentation task.

\section*{Acknowledgement}
The research was funded in part by C-BRIC, one of six centers in JUMP, a Semiconductor Research Corporation (SRC) program sponsored by DARPA, and the National Science Foundation (Grant\#1947826).


\bibliographystyle{stanfordcs}
\bibliography{ref}

\begin{thebibliography}{10}

\bibitem{he2016deep}
He, K., Zhang, X., Ren, S., Sun, J.:
\newblock Deep residual learning for image recognition.
\newblock (2016)  770--778

\bibitem{simonyan2014very}
Simonyan, K., Zisserman, A.:
\newblock Very deep convolutional networks for large-scale image recognition.
\newblock (2015)

\bibitem{girshick2015fast}
Girshick, R.:
\newblock Fast r-cnn.
\newblock In: Proceedings of the IEEE international conference on computer
  vision. (2015)  1440--1448

\bibitem{sze2017efficient}
Sze, V., Chen, Y.H., Yang, T.J., Emer, J.S.:
\newblock Efficient processing of deep neural networks: A tutorial and survey.
\newblock Proceedings of the IEEE \textbf{105}(12) (2017)  2295--2329

\bibitem{roy2019towards}
Roy, K., Jaiswal, A., Panda, P.:
\newblock Towards spike-based machine intelligence with neuromorphic computing.
\newblock Nature \textbf{575}(7784) (2019)  607--617

\bibitem{deng2020rethinking}
Deng, L., Wu, Y., Hu, X., Liang, L., Ding, Y., Li, G., Zhao, G., Li, P., Xie,
  Y.:
\newblock Rethinking the performance comparison between snns and anns.
\newblock Neural Networks \textbf{121} (2020)  294--307

\bibitem{furber2014spinnaker}
Furber, S.B., Galluppi, F., Temple, S., Plana, L.A.:
\newblock The spinnaker project.
\newblock Proceedings of the IEEE \textbf{102}(5) (2014)  652--665

\bibitem{akopyan2015truenorth}
Akopyan, F., Sawada, J., Cassidy, A., Alvarez-Icaza, R., Arthur, J., Merolla,
  P., Imam, N., Nakamura, Y., Datta, P., Nam, G.J.,  et~al.:
\newblock Truenorth: Design and tool flow of a 65 mw 1 million neuron
  programmable neurosynaptic chip.
\newblock IEEE transactions on computer-aided design of integrated circuits and
  systems \textbf{34}(10) (2015)  1537--1557

\bibitem{davies2018loihi}
Davies, M., Srinivasa, N., Lin, T.H., Chinya, G., Cao, Y., Choday, S.H., Dimou,
  G., Joshi, P., Imam, N., Jain, S.,  et~al.:
\newblock Loihi: A neuromorphic manycore processor with on-chip learning.
\newblock IEEE Micro \textbf{38}(1) (2018)  82--99

\bibitem{sengupta2019going}
Sengupta, A., Ye, Y., Wang, R., Liu, C., Roy, K.:
\newblock Going deeper in spiking neural networks: Vgg and residual
  architectures.
\newblock Frontiers in neuroscience \textbf{13} (2019) ~95

\bibitem{han2020rmp}
Han, B., Srinivasan, G., Roy, K.:
\newblock Rmp-snn: Residual membrane potential neuron for enabling deeper
  high-accuracy and low-latency spiking neural network.
\newblock In: Proceedings of the IEEE/CVF Conference on Computer Vision and
  Pattern Recognition. (2020)  13558--13567

\bibitem{diehl2015fast}
Diehl, P.U., Neil, D., Binas, J., Cook, M., Liu, S.C., Pfeiffer, M.:
\newblock Fast-classifying, high-accuracy spiking deep networks through weight
  and threshold balancing.
\newblock In: 2015 International Joint Conference on Neural Networks (IJCNN),
  ieee (2015)  1--8

\bibitem{rueckauer2017conversion}
Rueckauer, B., Lungu, I.A., Hu, Y., Pfeiffer, M., Liu, S.C.:
\newblock Conversion of continuous-valued deep networks to efficient
  event-driven networks for image classification.
\newblock Frontiers in neuroscience \textbf{11} (2017)  682

\bibitem{yurtsever2020survey}
Yurtsever, E., Lambert, J., Carballo, A., Takeda, K.:
\newblock A survey of autonomous driving: Common practices and emerging
  technologies.
\newblock IEEE Access \textbf{8} (2020)  58443--58469

\bibitem{treml2016speeding}
Treml, M., Arjona-Medina, J., Unterthiner, T., Durgesh, R., Friedmann, F.,
  Schuberth, P., Mayr, A., Heusel, M., Hofmarcher, M., Widrich, M.,  et~al.:
\newblock Speeding up semantic segmentation for autonomous driving.
\newblock In: MLITS, NIPS Workshop. Volume~2. (2016)

\bibitem{kim2019cnn}
Kim, Y., Kim, S., Kim, T., Kim, C.:
\newblock Cnn-based semantic segmentation using level set loss.
\newblock In: 2019 IEEE Winter Conference on Applications of Computer Vision
  (WACV), IEEE (2019)  1752--1760

\bibitem{zhang2018context}
Zhang, H., Dana, K., Shi, J., Zhang, Z., Wang, X., Tyagi, A., Agrawal, A.:
\newblock Context encoding for semantic segmentation.
\newblock In: Proceedings of the IEEE conference on Computer Vision and Pattern
  Recognition. (2018)  7151--7160

\bibitem{chen2018encoder}
Chen, L.C., Zhu, Y., Papandreou, G., Schroff, F., Adam, H.:
\newblock Encoder-decoder with atrous separable convolution for semantic image
  segmentation.
\newblock In: Proceedings of the European conference on computer vision (ECCV).
  (2018)  801--818

\bibitem{long2015fully}
Long, J., Shelhamer, E., Darrell, T.:
\newblock Fully convolutional networks for semantic segmentation.
\newblock In: Proceedings of the IEEE conference on computer vision and pattern
  recognition. (2015)  3431--3440

\bibitem{patrick2008128x}
Patrick, L., Posch, C., Delbruck, T.:
\newblock A 128x 128 120 db 15$\mu$ s latency asynchronous temporal contrast
  vision sensor.
\newblock IEEE journal of solid-state circuits \textbf{43} (2008)  566--576

\bibitem{lichtsteiner2006128}
Lichtsteiner, P., Posch, C., Delbruck, T.:
\newblock A 128 x 128 120db 30mw asynchronous vision sensor that responds to
  relative intensity change.
\newblock In: 2006 IEEE International Solid State Circuits Conference-Digest of
  Technical Papers, IEEE (2006)  2060--2069

\bibitem{posch2014retinomorphic}
Posch, C., Serrano-Gotarredona, T., Linares-Barranco, B., Delbruck, T.:
\newblock Retinomorphic event-based vision sensors: bioinspired cameras with
  spiking output.
\newblock Proceedings of the IEEE \textbf{102}(10) (2014)  1470--1484

\bibitem{delbruck2010activity}
Delbr{\"u}ck, T., Linares-Barranco, B., Culurciello, E., Posch, C.:
\newblock Activity-driven, event-based vision sensors.
\newblock In: Proceedings of 2010 IEEE International Symposium on Circuits and
  Systems, IEEE (2010)  2426--2429

\bibitem{neftci2019surrogate}
Neftci, E.O., Mostafa, H., Zenke, F.:
\newblock Surrogate gradient learning in spiking neural networks.
\newblock IEEE Signal Processing Magazine \textbf{36} (2019)  61--63

\bibitem{lee2016training}
Lee, J.H., Delbruck, T., Pfeiffer, M.:
\newblock Training deep spiking neural networks using backpropagation.
\newblock Frontiers in neuroscience \textbf{10} (2016)  508

\bibitem{wu2018spatio}
Wu, Y., Deng, L., Li, G., Zhu, J., Shi, L.:
\newblock Spatio-temporal backpropagation for training high-performance spiking
  neural networks.
\newblock Frontiers in neuroscience \textbf{12} (2018)  331

\bibitem{kim2020revisiting}
Kim, Y., Panda, P.:
\newblock Revisiting batch normalization for training low-latency deep spiking
  neural networks from scratch.
\newblock arXiv preprint arXiv:2010.01729 (2020)

\bibitem{panda2020toward}
Panda, P., Aketi, S.A., Roy, K.:
\newblock Toward scalable, efficient, and accurate deep spiking neural networks
  with backward residual connections, stochastic softmax, and hybridization.
\newblock Frontiers in Neuroscience \textbf{14} (2020)

\bibitem{cao2015spiking}
Cao, Y., Chen, Y., Khosla, D.:
\newblock Spiking deep convolutional neural networks for energy-efficient
  object recognition.
\newblock International Journal of Computer Vision \textbf{113}(1) (2015)
  54--66

\bibitem{diehl2015unsupervised}
Diehl, P.U., Cook, M.:
\newblock Unsupervised learning of digit recognition using
  spike-timing-dependent plasticity.
\newblock Frontiers in computational neuroscience \textbf{9} (2015) ~99

\bibitem{comsa2020temporal}
Comsa, I.M., Fischbacher, T., Potempa, K., Gesmundo, A., Versari, L.,
  Alakuijala, J.:
\newblock Temporal coding in spiking neural networks with alpha synaptic
  function.
\newblock In: ICASSP 2020-2020 IEEE International Conference on Acoustics,
  Speech and Signal Processing (ICASSP), IEEE (2020)  8529--8533

\bibitem{venkatesha2021federated}
Venkatesha, Y., Kim, Y., Tassiulas, L., Panda, P.:
\newblock Federated learning with spiking neural networks.
\newblock arXiv preprint arXiv:2106.06579 (2021)

\bibitem{kim2021optimizing}
Kim, Y., Panda, P.:
\newblock Optimizing deeper spiking neural networks for dynamic vision sensing.
\newblock Neural Networks (2021)

\bibitem{kim2021privatesnn}
Kim, Y., Venkatesha, Y., Panda, P.:
\newblock Privatesnn: Fully privacy-preserving spiking neural networks.
\newblock arXiv preprint arXiv:2104.03414 (2021)

\bibitem{mostafa2017supervised}
Mostafa, H.:
\newblock Supervised learning based on temporal coding in spiking neural
  networks.
\newblock IEEE transactions on neural networks and learning systems
  \textbf{29}(7) (2017)  3227--3235

\bibitem{park2020t2fsnn}
Park, S., Kim, S., Na, B., Yoon, S.:
\newblock T2fsnn: Deep spiking neural networks with time-to-first-spike coding.
\newblock arXiv preprint arXiv:2003.11741 (2020)

\bibitem{gerstner2002spiking}
Gerstner, W., Kistler, W.M.:
\newblock Spiking neuron models: Single neurons, populations, plasticity.
\newblock Cambridge university press (2002)

\bibitem{bi1998synaptic}
Bi, G.q., Poo, M.m.:
\newblock Synaptic modifications in cultured hippocampal neurons: dependence on
  spike timing, synaptic strength, and postsynaptic cell type.
\newblock Journal of neuroscience \textbf{18}(24) (1998)  10464--10472

\bibitem{hebb2005organization}
Hebb, D.O.:
\newblock The organization of behavior: A neuropsychological theory.
\newblock Psychology Press (2005)

\bibitem{bliss1993synaptic}
Bliss, T.V., Collingridge, G.L.:
\newblock A synaptic model of memory: long-term potentiation in the
  hippocampus.
\newblock Nature \textbf{361}(6407) (1993)  31--39

\bibitem{yousefzadeh2018practical}
Yousefzadeh, A., Stromatias, E., Soto, M., Serrano-Gotarredona, T.,
  Linares-Barranco, B.:
\newblock On practical issues for stochastic stdp hardware with 1-bit synaptic
  weights.
\newblock Frontiers in neuroscience \textbf{12} (2018)  665

\bibitem{jin2010implementing}
Jin, X., Rast, A., Galluppi, F., Davies, S., Furber, S.:
\newblock Implementing spike-timing-dependent plasticity on spinnaker
  neuromorphic hardware.
\newblock In: The 2010 International Joint Conference on Neural Networks
  (IJCNN), IEEE (2010)  1--8

\bibitem{kim2021visual}
Kim, Y., Panda, P.:
\newblock Visual explanations from spiking neural networks using interspike
  intervals.
\newblock arXiv preprint arXiv:2103.14441 (2021)

\bibitem{fang2019swarm}
Fang, Y., Wang, Z., Gomez, J., Datta, S., Khan, A.I., Raychowdhury, A.:
\newblock A swarm optimization solver based on ferroelectric spiking neural
  networks.
\newblock Frontiers in neuroscience \textbf{13} (2019)  855

\bibitem{frady2020neuromorphic}
Frady, E.P., Orchard, G., Florey, D., Imam, N., Liu, R., Mishra, J., Tse, J.,
  Wild, A., Sommer, F.T., Davies, M.:
\newblock Neuromorphic nearest neighbor search using intel's pohoiki springs.
\newblock In: Proceedings of the Neuro-inspired Computational Elements
  Workshop. (2020)  1--10

\bibitem{kim2020spiking}
Kim, S., Park, S., Na, B., Yoon, S.:
\newblock Spiking-yolo: spiking neural network for energy-efficient object
  detection.
\newblock In: Proceedings of the AAAI Conference on Artificial Intelligence.
  Volume~34. (2020)  11270--11277

\bibitem{ronneberger2015u}
Ronneberger, O., Fischer, P., Brox, T.:
\newblock U-net: Convolutional networks for biomedical image segmentation.
\newblock In: International Conference on Medical image computing and
  computer-assisted intervention, Springer (2015)  234--241

\bibitem{cciccek20163d}
{\c{C}}i{\c{c}}ek, {\"O}., Abdulkadir, A., Lienkamp, S.S., Brox, T.,
  Ronneberger, O.:
\newblock 3d u-net: learning dense volumetric segmentation from sparse
  annotation.
\newblock In: International conference on medical image computing and
  computer-assisted intervention, Springer (2016)  424--432

\bibitem{janai2020computer}
Janai, J., G{\"u}ney, F., Behl, A., Geiger, A.,  et~al.:
\newblock Computer vision for autonomous vehicles: Problems, datasets and state
  of the art.
\newblock Foundations and Trends{\textregistered} in Computer Graphics and
  Vision \textbf{12}(1--3) (2020)  1--308

\bibitem{minaee2020image}
Minaee, S., Boykov, Y., Porikli, F., Plaza, A., Kehtarnavaz, N., Terzopoulos,
  D.:
\newblock Image segmentation using deep learning: A survey.
\newblock arXiv preprint arXiv:2001.05566 (2020)

\bibitem{noh2015learning}
Noh, H., Hong, S., Han, B.:
\newblock Learning deconvolution network for semantic segmentation.
\newblock In: Proceedings of the IEEE international conference on computer
  vision. (2015)  1520--1528

\bibitem{lin2017refinenet}
Lin, G., Milan, A., Shen, C., Reid, I.:
\newblock Refinenet: Multi-path refinement networks for high-resolution
  semantic segmentation.
\newblock In: Proceedings of the IEEE conference on computer vision and pattern
  recognition. (2017)  1925--1934

\bibitem{tensorflow2015-whitepaper}
Abadi, M., Agarwal, A., Barham, P., Brevdo, E., Chen, Z., Citro, C., Corrado,
  G.S., Davis, A., Dean, J., Devin, M., Ghemawat, S., Goodfellow, I., Harp, A.,
  Irving, G., Isard, M., Jia, Y., Jozefowicz, R., Kaiser, L., Kudlur, M.,
  Levenberg, J., Man\'{e}, D., Monga, R., Moore, S., Murray, D., Olah, C.,
  Schuster, M., Shlens, J., Steiner, B., Sutskever, I., Talwar, K., Tucker, P.,
  Vanhoucke, V., Vasudevan, V., Vi\'{e}gas, F., Vinyals, O., Warden, P.,
  Wattenberg, M., Wicke, M., Yu, Y., Zheng, X.:
\newblock {TensorFlow}: Large-scale machine learning on heterogeneous systems
  (2015) Software available from tensorflow.org.

\bibitem{meftah2010segmentation}
Meftah, B., Lezoray, O., Benyettou, A.:
\newblock Segmentation and edge detection based on spiking neural network
  model.
\newblock Neural Processing Letters \textbf{32}(2) (2010)  131--146

\bibitem{zheng2019image}
Zheng, D., Lin, X., Wang, X.:
\newblock Image segmentation method based on spiking neural network with
  adaptive synaptic weights.
\newblock In: 2019 IEEE 4th International Conference on Signal and Image
  Processing (ICSIP), IEEE (2019)  1043--1049

\bibitem{chen2017deeplab}
Chen, L.C., Papandreou, G., Kokkinos, I., Murphy, K., Yuille, A.L.:
\newblock Deeplab: Semantic image segmentation with deep convolutional nets,
  atrous convolution, and fully connected crfs.
\newblock IEEE transactions on pattern analysis and machine intelligence
  \textbf{40}(4) (2017)  834--848

\bibitem{everingham2010pascal}
Everingham, M., Van~Gool, L., Williams, C.K., Winn, J., Zisserman, A.:
\newblock The pascal visual object classes (voc) challenge.
\newblock International journal of computer vision \textbf{88}(2) (2010)
  303--338

\bibitem{kim2018deep}
Kim, J., Kim, H., Huh, S., Lee, J., Choi, K.:
\newblock Deep neural networks with weighted spikes.
\newblock Neurocomputing \textbf{311} (2018)  373--386

\bibitem{hariharan2011semantic}
Hariharan, B., Arbel{\'a}ez, P., Bourdev, L., Maji, S., Malik, J.:
\newblock Semantic contours from inverse detectors.
\newblock In: 2011 International Conference on Computer Vision, IEEE (2011)
  991--998

\bibitem{binas2017ddd17}
Binas, J., Neil, D., Liu, S.C., Delbruck, T.:
\newblock Ddd17: End-to-end davis driving dataset.
\newblock arXiv preprint arXiv:1711.01458 (2017)

\bibitem{alonso2019ev}
Alonso, I., Murillo, A.C.:
\newblock Ev-segnet: Semantic segmentation for event-based cameras.
\newblock In: Proceedings of the IEEE/CVF Conference on Computer Vision and
  Pattern Recognition Workshops. (2019)  0--0

\bibitem{ioffe2015batch}
Ioffe, S., Szegedy, C.:
\newblock Batch normalization: Accelerating deep network training by reducing
  internal covariate shift.
\newblock arXiv preprint arXiv:1502.03167 (2015)

\bibitem{rathi2020diet}
Rathi, N., Roy, K.:
\newblock Diet-snn: Direct input encoding with leakage and threshold
  optimization in deep spiking neural networks.
\newblock arXiv preprint arXiv:2008.03658 (2020)

\bibitem{horowitz20141}
Horowitz, M.:
\newblock 1.1 computing's energy problem (and what we can do about it).
\newblock In: 2014 IEEE International Solid-State Circuits Conference Digest of
  Technical Papers (ISSCC), IEEE (2014)  10--14

\end{thebibliography}

\end{document}